\pdfoutput=1
\documentclass[11pt]{article}

\usepackage[final]{EMNLP2023}

\usepackage{times}
\usepackage{latexsym}
\usepackage{booktabs}
\usepackage{multirow} 
\usepackage{graphicx}
\usepackage{subcaption}
\usepackage{makecell}  % For multiline cells
\usepackage{xcolor}
\usepackage{float} 
\usepackage{footmisc}
\usepackage[most]{tcolorbox}
\usepackage{lipsum} % Optional, just for demo text
\tcbuselibrary{breakable}
\usepackage{listings}
\usepackage{enumitem}

\usepackage[table]{xcolor}
\definecolor{catblue}{HTML}{B7CDE8}    % for (a)
\definecolor{catyellow}{HTML}{F7DE9F}  % for (b)
\definecolor{catgreen}{HTML}{C9E3C5}   % for (c)

\usepackage[T1]{fontenc}
\usepackage[utf8]{inputenc}

\usepackage{microtype}

\usepackage{inconsolata}

\title{SciEvent: Benchmarking Multi-domain Scientific Event Extraction}

\author{
Bofu Dong\textsuperscript{1}, Pritesh Shah\textsuperscript{1}, Sumedh Sonawane\textsuperscript{1}, Tiyasha Banerjee\textsuperscript{1}, Erin Brady\textsuperscript{1}\\
\textbf{Xinya Du\textsuperscript{2}, Ming Jiang\textsuperscript{1,3}} \\
\normalsize \textsuperscript{1}Indiana University Indianapolis, 
\normalsize \textsuperscript{2}University of Texas at Dallas, 
\normalsize \textsuperscript{3}University of Wisconsin-Madison \\
\normalsize \texttt{bofudong@iu.edu} \qquad \normalsize \texttt{ming.jiang@wisc.edu}
}
\begin{document}
\maketitle

\begin{abstract}
Scientific information extraction (SciIE) has primarily relied on entity-relation extraction in narrow domains, limiting its applicability to interdisciplinary research and struggling to capture the necessary context of scientific information, often resulting in fragmented or conflicting statements. In this paper, we introduce SciEvent\footnote{Our code and benchmark are released at https://github.com/desdai/SciEvent.}, a novel multi-domain benchmark of scientific abstracts annotated via a unified event extraction (EE) schema designed to enable structured and context-aware understanding of scientific content. It includes 500 abstracts across five research domains, with manual annotations of event segments, triggers, and fine-grained arguments. We define SciIE as a multi-stage EE pipeline: (1) segmenting abstracts into core scientific activities---\textit{Background}, \textit{Method}, \textit{Result}, and \textit{Conclusion}; and (2) extracting the corresponding triggers and arguments. Experiments with fine-tuned EE models, large language models (LLMs), and human annotators reveal a performance gap, with current models struggling in domains such as sociology and humanities. SciEvent serves as a challenging benchmark and a step toward generalizable, multi-domain SciIE.
\end{abstract}
\section{Introduction}
\label{sec:introduction}
Scientific information extraction (SciIE) distills structured knowledge from unstructured scientific articles and supports key scientific applications such as literature review \cite{hong2021challenges}, paper recommendation \cite{ikoma-matsubara-2023-paper}, and knowledge discovery \cite{stavropoulos-etal-2023-empowering}, especially in recent years as many domains are facing a publication deluge. 

\begin{figure}[t]
    \centering \includegraphics[width=0.98\linewidth]{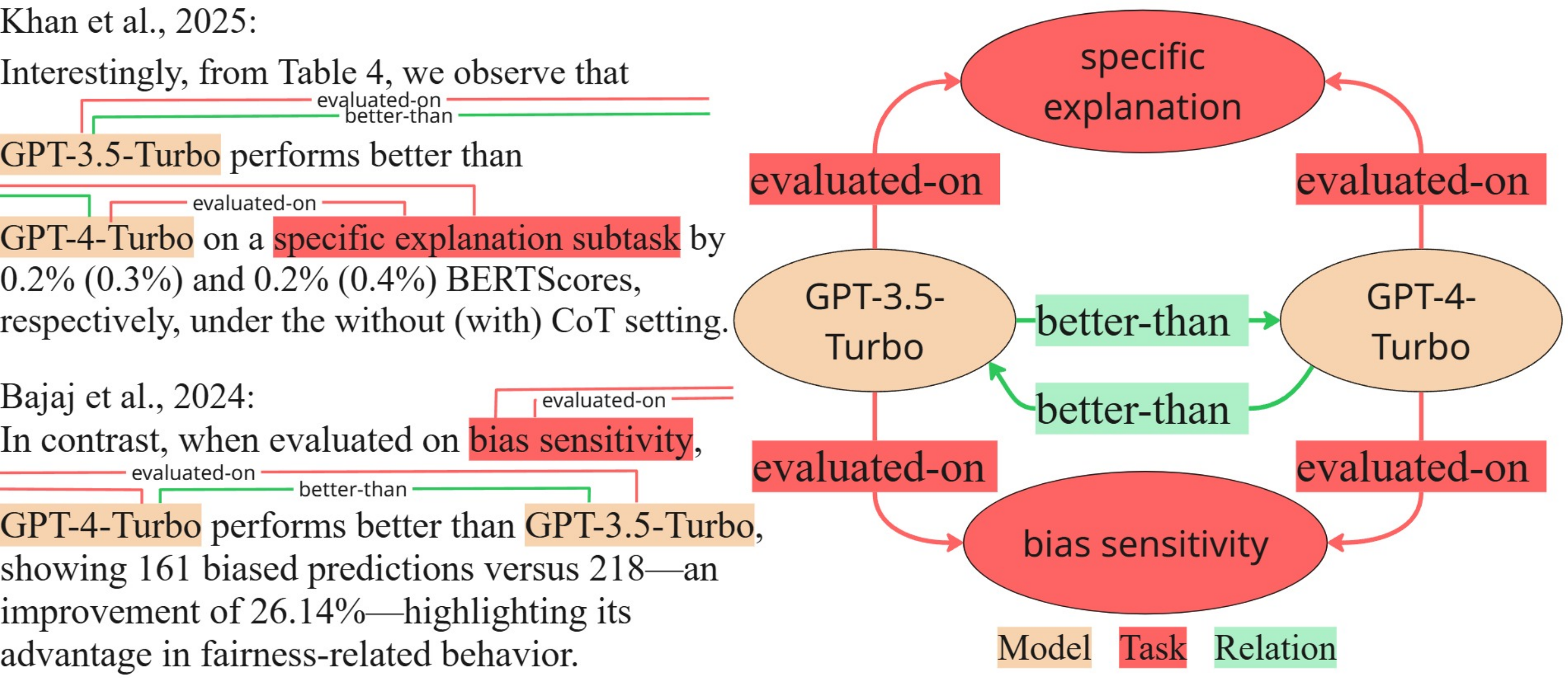}
    \caption{Conflicting statements in entity-relation extraction. $\langle$GPT-3.5-Turbo, better than, GPT-4-Turbo$\rangle$ vs. $\langle$GPT-4-Turbo, better than, GPT-3.5-Turbo$\rangle$}
    \label{fig: ere not good}
    \vspace{-3mm}
\end{figure}

Existing works on SciIE generally follow an entity-relation extraction (ERE) paradigm that aims to extract isolated scientific concepts and connect them by identifying semantic relations, either binary \citep{luan-etal-2018-multi, zhang-etal-2024-scier} or $N$-ary (\citealp{jain-etal-2020-scirex}; \citealp{zhuang-etal-2022-resel}). Despite remarkable contributions made by prior studies, one major concern is that representing scientific content as disconnected entity-relation tuples may fragment the underlying narrative and even introduce conflicting statements, especially when synthesizing information across multiple publications. As shown in Figure~\ref{fig: ere not good}, one paper may generate the tuple $\langle$ ``GPT-3.5-Turbo'', ``better than'', ``GPT-4-Turbo''$\rangle$, while another produces the opposite. Lacking contextual cues such as task setup or evaluation criteria, these tuples alone fail to convey meaningful or reliable scientific insights.

Inspired by the heavily context-dependent nature of scientific publications, we adopt an event extraction (EE) paradigm. This paradigm focuses on identifying triggers that best represent each event and extracting associated arguments, which are then assigned specific semantic roles. This enables a more structured and context-aware representation of important scientific information. Despite its potential for representing scientific information, a major limitation of existing EE efforts in the scientific domain is their narrow focus on specific fields, often resulting in the development of domain-specific EE schemas. For example, \citet{zhang-etal-2024-scier} and \citet{jain-etal-2020-scirex} focus on machine learning, and \citet{kim-etal-2011-overview-genia} focus on bio-molecule area. Given the rapid growth of interdisciplinary research in recent years (\citealp{leto-etal-2024-first}; \citealp{okamura2019interdisciplinarity}), there is an increasing need for a unified scientific EE schema capable of generalizing across diverse scholarly domains.

To address this gap, we introduce SciEvent, a unified EE schema for scientific texts, along with a dataset featuring manually annotated events and fine-grained arguments drawn from diverse research abstracts. Building on this dataset, we define three SciIE tasks: (1) event segmentation, which involves dividing the text into spans that represent core scientific activities such as \textit{Background}, \textit{Method}, \textit{Result}, and \textit{Conclusion}; (2) trigger identification, which aims to detect the key anchor of each scientific event; and (3) argument extraction, which focuses on identifying the arguments involved in each scientific activity and assigning them roles such as context, method, or result. Differing from conventional EE pipelines, we introduce event segmentation as a preliminary task, recognizing that events in scientific texts often span multiple sentences and lack clear boundaries. Additionally, trigger words in scientific texts—such as ``show'', ``demonstrate'', or ``present''—are frequently shared across different event types. Without first segmenting the text into discrete events, it becomes challenging to accurately delineate event boundaries, increasing the risk of misinterpreting or misclassifying both triggers and their associated arguments.

SciEvent contains 500 abstracts from five diverse scientific domains, each fully annotated using an EE paradigm. To evaluate the challenges posed by this dataset, we assess the performance of fine-tuned EE models, tuning-free large language models (LLMs), and human annotators. The results demonstrate SciEvent’s broad domain coverage and reveal that existing models consistently lag behind human performance. This gap highlights the limitations of current approaches and the absence of EE models capable of generalizing across scientific domains.
\section{Related Work}
\label{sec:related}

\paragraph{Event Extraction}
Existing work on event extraction (EE) typically frames the task via two paradigms. One is trigger-argument extraction (\citealp{LDC2006T06}; \citealp{hsu-etal-2022-degree}; \citealp{lin-etal-2020-joint}), where the trigger serves as the event anchor, most clearly signaling the occurrence of an event, while the arguments represent entity mentions that participate in the event, each fulfilling distinct roles. The other one treats EE as a trigger-free template-filling task (\citealp{muc-1992-message}; \citealp{du-cardie-2020-document}; \citealp{huang-etal-2021-document}), aiming to extract event-relevant arguments and assigning them to specific roles within each event template. The latter mainly focuses on document-level EE \cite{du-cardie-2020-document}, while the former has been widely used in both sentence-level \cite{LDC2006T06} and document-level EE \cite{li-etal-2021-document}. Our benchmark follows the trigger-argument paradigm.

Regarding EE benchmarks, prior studies have largely focused on data in generic domains. Popular examples include newswire (\citealp{grishman-sundheim-1996-message}; \citealp{nguyen-etal-2016-dataset}; \citealp{doddington-etal-2004-automatic}; \citealp{ebner-etal-2020-multi}; \citealp{song-etal-2015-light}), Wikipedia (\citealp{li-etal-2021-document}; \citealp{pouran-ben-veyseh-etal-2022-mee}), social media (\citealp{sharif-etal-2024-explicit}; \citealp{10.5555/3171837.3171867}; \citealp{10.1145/3332185}) and widely-used knowledgebases like FrameNet \cite{baker-etal-1998-berkeley-framenet} and PropBank \cite{bonial-etal-2014-propbank}. While some researchers have broadened the scope of EE to scientific literature, their efforts tend to center the biomedical domain, particularly emphasizing state changes and interactions between biomolecules such as genes and proteins (\citealp{kim-etal-2011-overview-genia}; \citealp{10.1093/bioinformatics/bts407}; \citealp{kim-etal-2013-genia}). Differing from prior work, we extend EE to encompass a broader range of scientific domains, creating a unified annotation schema designed to facilitate interdisciplinary information extraction.
\vspace{-2mm}
\paragraph{Scientific Information Extraction} Research on scientific information extraction (IE) primarily targets two main types of information: (1) citation-based analysis, which involves identifying either binary citation influence classification (\citealp{kunnath-etal-2020-overview}; \citealp{n-kunnath-etal-2021-overview}; \citealp{maheshwari-etal-2021-scibert}) or multi-class citation intents (purpose) classification (\citealp{cohan-etal-2019-structural}; \citealp{jurgens-etal-2018-measuring}), and (2) content-based analysis (\citealp{gupta-manning-2011-analyzing}; \citealp{10.1145/2505515.2505613}; \citealp{gabor2016unsupervised}; \citealp{pronesti2025querydrivendocumentlevelscientificevidence}), which primarily focuses on extracting scientific entities, supporting evidence, and semantic relationships among them, with the ultimate goal of building concept-centric knowledge graphs (\citealp{ma-etal-2022-mmekg}; \citealp{10.1145/3366423.3380107}; \citealp{10.1609/aaai.v33i01.33013027}). For example, SciERC \cite{luan-etal-2018-multi}, consists of 500 scientific abstracts annotated with scientific entities, their pairwise relations, and coreference clusters. SciREX \cite{jain-etal-2020-scirex} provides annotations across 438 full documents, covering four entity types: TASK, DATASET, METHOD, and METRIC. Beyond general knowledge extraction, some studies further focus on specific research subjects. This line of work designs domain-specific event extraction tasks to capture fine-grained scientific activities (\citealp{he-etal-2024-zsee},\citealp{kim-etal-2011-overview-genia}, \citealp{huang-etal-2020-biomedical}, \citealp{bjorne-etal-2010-scaling}). For example, various biomedical EE tasks have been proposed to investigate biological processes such as protein-protein and gene-disease interactions (\citealp{kim-etal-2013-genia}; \citealp{kim-etal-2011-overview-genia}; \citealp{bjorne-etal-2010-scaling}). Our work similarly focuses on scientific EE. However, differing from prior works targeting specific domain, we aim to design a unified schema for organizing general scientific activities across diverse scientific domains.
\section{SciEvent Benchmark}
\label{sec:dataset}
\vspace{-2mm}
\paragraph{Data Collection}
To support cross-domain evaluation and capture diverse writing conventions, we select publicly available, peer-reviewed scientific abstracts published in 2023 to reflect contemporary language use.
We select five domains: natural language processing (NLP) from the Annual Meeting of the Association for Computational Linguistics (ACL) \citep{ACL2023}, 
social computing (SC) from the Proceedings of the ACM on Human-Computer Interaction (CSCW) \citep{CSCW2023}, 
medical informatics (MI) from the Journal of Medical Internet Research (JMIR) \citep{JMIR2023}, 
computational biology (CB) from the Bioinformatics \citep{Bioinfo2023}, 
and digital humanities (DH) from the Digital Humanities Quarterly \citep{DHQ}\footnote{Full source attributions are included in the benchmark metadata available in our public GitHub repository.}.

These domains are selected for their methodological diversity, resource availability, relevance to interdisciplinary research, and representativeness of their respective fields. NLP and CB domains are well-studied and offer structured, technical abstracts, while SC and DH are underrepresented and characterized by more narrative, context-rich writing.
To support document-level modeling, we retain abstracts with at least three sentences and two identifiable events, filtering out those that are too short to provide meaningful structure.
In total, we collect 500 scientific abstracts---100 each in NLP, SC, and CB, 120 in DH, and 80 in MI.
DH has fewer publications and shorter abstracts, so we extend the sampling range to 2021–-2023 and include 120 abstracts to ensure sufficient coverage of domain variation. MI abstracts are longer and denser, so we select 80 abstracts to balance event content comparability across all five domains. We conduct a detailed keyword analysis based on each domain’s call for papers to ensure comprehensive coverage and minimize bias in our dataset. Additional details are provided in Appendix~\ref{app:keyword}

\vspace{-2mm}
\paragraph{Annotation Pipeline}
Overall, our annotation pipeline consists of two stages: (1) event segmentation, and (2) trigger-argument extraction. 
In the first stage, we segment an abstract into four event types: \textit{Background}, \textit{Method}, \textit{Result}, and \textit{Conclusion}, which are adopted from the most common aspects of scientific publications \citep{ripple2012structured}. In the second stage, we further annotate each segment at a fine-grained level,  focusing on identifying the event trigger and role-specific arguments.

In prior event extraction works, particularly in newswire and broadcast domains, triggers like ``attack'' define clear and stable event frames, with roles such as ``attacker'' and ``target'' naturally grounded in the trigger’s semantics. In scientific texts, however, single-word triggers like ``show'' lack this clarity. Even after event segmentation and the event type (e.g., ``Result'') is known, the trigger alone does not specify what the event is about. Roles like ``people who show'' or ``shown item'' are not meaningful on their own, as the event’s meaning depends on the full proposition. For example, ``showing a promising result'' differs from ``showing a methodological limitation''. With this consideration, we represent the trigger as a tuple of $\langle$Agent, Action, Object$\rangle$, anchoring the event in its core semantics. Notably, our empirical investigation on raw data shows that in some cases, the object in an event trigger may consist of two non-contiguous text spans. For example, ``protein sequences'' and ``gene expression profiles'' in a ``Method'' event: ``We analyze protein sequences, which exhibit structural variation, and gene expression profiles \dots'' are the objects of the action ``analyze''. Accordingly, we specify the labels \textit{Primary Object} and \textit{Secondary Object} for annotation. When we have two annotated object spans in an event, we concatenated them for further analysis.

Given the trigger identified per event, we then annotate its relevant arguments. We define nine argument roles: \textit{Context}, \textit{Purpose}, \textit{Method}, \textit{Result}, \textit{Analysis}, \textit{Challenge}, \textit{Ethical}, \textit{Implication}, and \textit{Contradiction}. Each role targets a specific dimension of scientific abstract, adapted from Core Scientific Concepts \cite{Liakata2012ConceptualizationZones} and inspired by scientific writing guides (\citealp{Paltridge2002}; \citealp{Alley1996}). While some argument roles share names with event types (e.g., Method, Result), they are not restricted to those events. For example, evaluation method often appear within the Result event. We attach the detailed codebook in Appendix~\ref{app:codebook}.

\vspace{-2mm}
\paragraph{Annotation Quality} We employ five graduate students as annotators, all specializing in NLP and are either native English speakers or PhD students. Each annotator has domain expertise in at least one of the five selected fields, ensuring comprehensive coverage across all diverse scientific areas. To evaluate the quality of event segment annotations, we randomly sample 10 abstracts per domain and had two annotators independently annotate each. Intercoder reliability, measured by Cohen’s Kappa \cite{Cohen1960ACO}, is $0.83$, showing strong agreement.

Considering that trigger and argument extraction involve more fine-grained and complex annotations than event segmentation, it increases the likelihood of annotator disagreement. To ensure consistency, we employ a collaborative, multi-round, discussion-based annotation process \cite{oortwijn-etal-2021-interrater} rather than a single-pass approach. Annotators first label the data independently, followed by review sessions with a meta-annotator to enforce codebook alignment. This cycle is repeated over six rounds, yielding 100\% agreement on triggers and 95.41\% agreement (4703/4929) on arguments. The remaining 4.59\% are resolved through majority voting among all five annotators, resulting in full team consensus on the final annotations. Notably, all disagreements in trigger and argument annotations, such as span variations and ambiguous argument roles, are resolved through specific rules outlined in the Codebook (Appendix~\ref{app: Annotation Rule}). To assess human performance for comparison with models, we additionally recruit six untrained annotators to independently annotate a randomly selected subset of our benchmark (consisting of 75 abstracts). For these annotators, we only provide them with a brief task description and basic instructions for using the annotation interface (see appendix~\ref{app:tool}).
\vspace{-3mm}
\begin{figure}[h]
  \centering
  \includegraphics[width=0.98\linewidth]{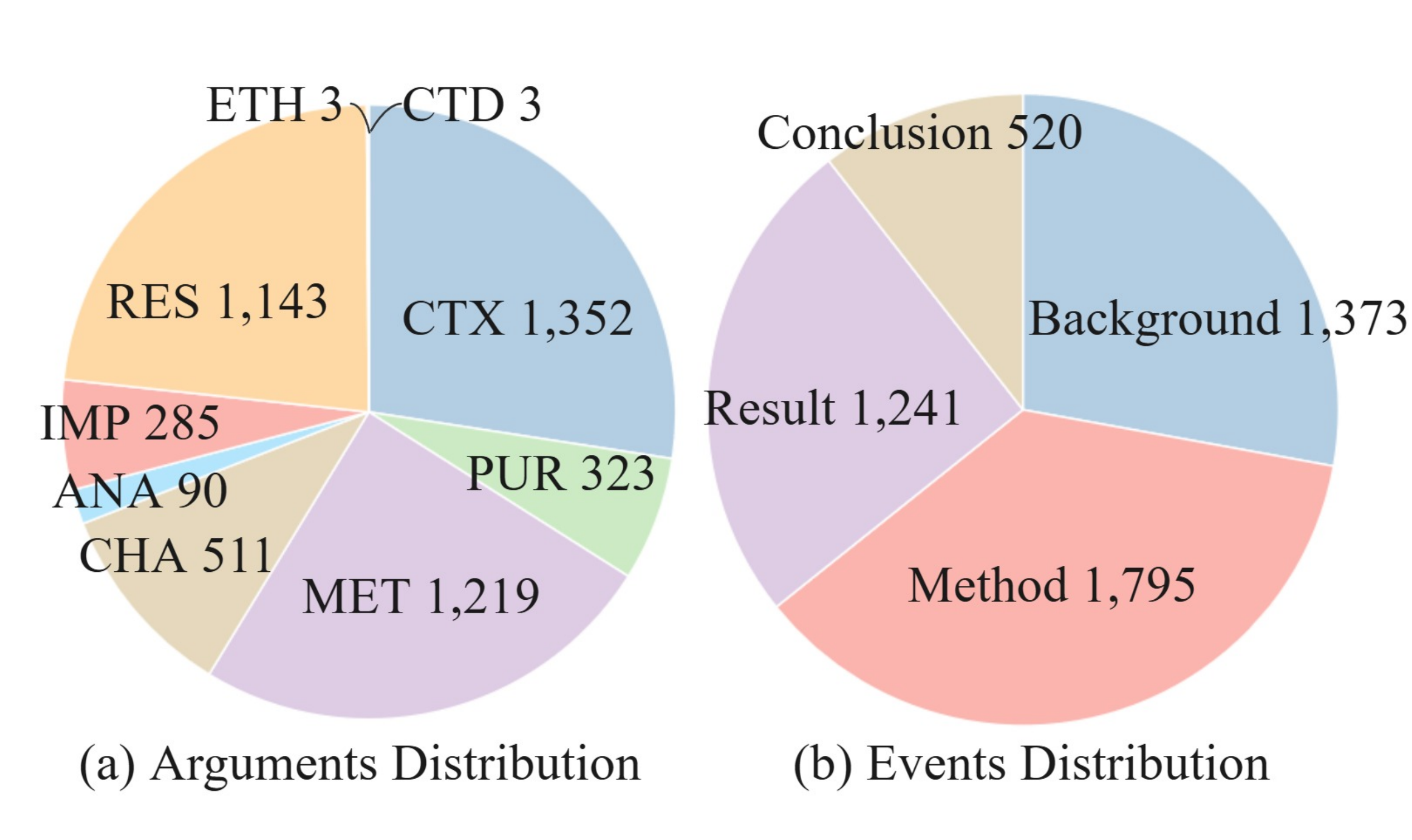}
  \caption{Distribution of (a) argument roles and (b) event types across the dataset.}
  \label{fig:piechart2}
  \vspace{-3mm}
\end{figure}
\paragraph{Data Analysis}
% \usepackage{booktabs} % Add to preamble

% \begin{table*}[h]
% \centering
% \small
% % \renewcommand{\arraystretch}{1.1}
% \setlength{\tabcolsep}{1.5pt}
% \resizebox{\columnwidth}{!}{%
% \begin{tabular}{@{}llllll@{}}
% \toprule
% \textbf{Dataset} & \textbf{\#Doc} & \textbf{\#Mentions} & \textbf{Paradigm} & \textbf{Source} & \textbf{Domains} \\
% \midrule
% \textsc{SciREX} & 438 & 8,592 & ERE & Full paper & ML \\
% \textsc{SciERC} & 500 & 8,089 & ERE & Abstract & Speech, ML, CV, AI \\
% \textsc{SemEval17} & 493 & 8,529 & ERE & Paragraph & CS, MS, Physics \\
% \textsc{SemEval18} & 500 & 7,505 & ERE & Abstract & CL \\
% \textsc{SciER} & 106 & 24,518 & ERE & Full paper & ML \\
% \textsc{Genia2011} & 1,224 & 21,549 & EE & Abstract + Full & BioMol \\
% \hline
% \textsc{SciEvent (ours)} & 500 & 8,911 & EE & Abstract & NLP, SC, CB, MI, DH \\
% \bottomrule
% \end{tabular}
% }
% \caption{Comparison of scientific IE datasets. Abbreviations: \textbf{NLP} = Natural Language Processing, \textbf{SC} = Social Computing, \textbf{CB} = Computational Biology, \textbf{MI} = Medical Informatics, \textbf{DH} = Digital Humanities, \textbf{ML} = Machine Learning, \textbf{AI} = Artificial Intelligence, \textbf{CV} = Computer Vision, \textbf{CS} = Computer Science, \textbf{MS} = Material Science, \textbf{CL} = Computational Linguistics, \textbf{BioMol} = Biomolecular.}
% \label{tab:dataset-comparison}
% \vspace{-3mm}
% \end{table*}

\begin{table*}[h]
\centering
\small
\setlength{\tabcolsep}{3pt}
\begin{tabular}{@{}lcccclll@{}}
\toprule
\textbf{Dataset} & \textbf{\#Doc} & \textbf{\#Mentions} & \textbf{Arg./Ent. Types} & \textbf{Avg Sent./Evt} & \textbf{Paradigm} & \textbf{Source} & \textbf{Domains} \\
\midrule
\textsc{SciREX} & 438 & 8,592 & 4 & - & ERE & Full paper & ML \\
\textsc{SciERC} & 500 & 8,089 & 6 & - & ERE & Abstract & Speech, ML, CV, AI \\
\textsc{SemEval17} & 493 & 8,529 & 3 & - & ERE & Paragraph & CS, MS, Physics \\
\textsc{SemEval18} & 500 & 7,505 & 1 & - & ERE & Abstract & CL \\
\textsc{SciER} & 106 & 24,518 & 3 & - & ERE & Full paper & ML \\
\textsc{Genia2011} & 1,224 & 21,549 & 10 & 1 & EE & Abstract/Full & BioMol \\
\hline
\textsc{SciEvent (ours)} & 500 & 8,911 & 9 & 2.95 & EE & Abstract & NLP, SC, CB, MI, DH \\
\bottomrule
\end{tabular}
\caption{Comparison of scientific IE datasets. Abbreviations: \textbf{Arg./Ent. Types} = Argument/Entity Types, \textbf{Avg Sent./Evt} = Average Sentence Per Event, \textbf{NLP} = Natural Language Processing, \textbf{SC} = Social Computing, \textbf{CB} = Computational Biology, \textbf{MI} = Medical Informatics, \textbf{DH} = Digital Humanities, \textbf{ML} = Machine Learning, \textbf{AI} = Artificial Intelligence, \textbf{CV} = Computer Vision, \textbf{CS} = Computer Science, \textbf{MS} = Material Science, \textbf{CL} = Computational Linguistics, \textbf{BioMol} = Biomolecular.}
\label{tab:dataset-comparison}
\vspace{-3mm}
\end{table*}

% % \usepackage{booktabs} % Add to preamble

% \begin{table*}[h]
% \centering
% \small
% % \renewcommand{\arraystretch}{1.1}
% \setlength{\tabcolsep}{1.5pt}
% \begin{tabular}{@{}lcccccll@{}}
% \toprule
% \textbf{Dataset} & \textbf{\#Doc} & \textbf{\#Mentions} & \textbf{Event} & \textbf{Arg./Ent.} & \textbf{Avg Sent./Evt} & \textbf{Source} & \textbf{Domains} \\
% \midrule
% \textsc{SciREX} & 438 & 8,592 & - & 4 & - & Full paper & ML \\
% \textsc{SciERC} & 500 & 8,089 & - & 6 & - & Abstract & Speech, ML, CV, AI \\
% \textsc{SemEval17} & 493 & 8,529 & - & 3 & - & Paragraph & CS, MS, Physics \\
% \textsc{SemEval18} & 500 & 7,505 & - & 1 & - & Abstract & CL \\
% \textsc{SciER} & 106 & 24,518 & - & 3 & - & Full paper & ML \\
% \textsc{Genia2011} & 1,224 & 21,549 & 9 & 10 & 1 & Abstract + Full & BioMol \\
% \hline
% \textsc{SciEvent (ours)} & 500 & 8,911 & 4 & 9 & 2.95 & Abstract & NLP, SC, CB, MI, DH \\
% \bottomrule
% \end{tabular}
% \caption{Comparison of scientific IE datasets. Abbreviations: \textbf{NLP} = Natural Language Processing, \textbf{SC} = Social Computing, \textbf{CB} = Computational Biology, \textbf{MI} = Medical Informatics, \textbf{DH} = Digital Humanities, \textbf{ML} = Machine Learning, \textbf{AI} = Artificial Intelligence, \textbf{CV} = Computer Vision, \textbf{CS} = Computer Science, \textbf{MS} = Material Science, \textbf{CL} = Computational Linguistics, \textbf{BioMol} = Biomolecular.}
% \label{tab:dataset-comparison}
% \vspace{-3mm}
% \end{table*}
Using the above annotation pipeline, we construct a dataset of 500 annotated scientific abstracts containing 8,911 structured mentions, as shown in Table~\ref{tab:dataset-comparison}. Its broad domain coverage supports robust cross-domain analysis.

\begin{figure}[h]
  \centering
  % spans both columns at full text width
  \includegraphics[width=0.98\linewidth]{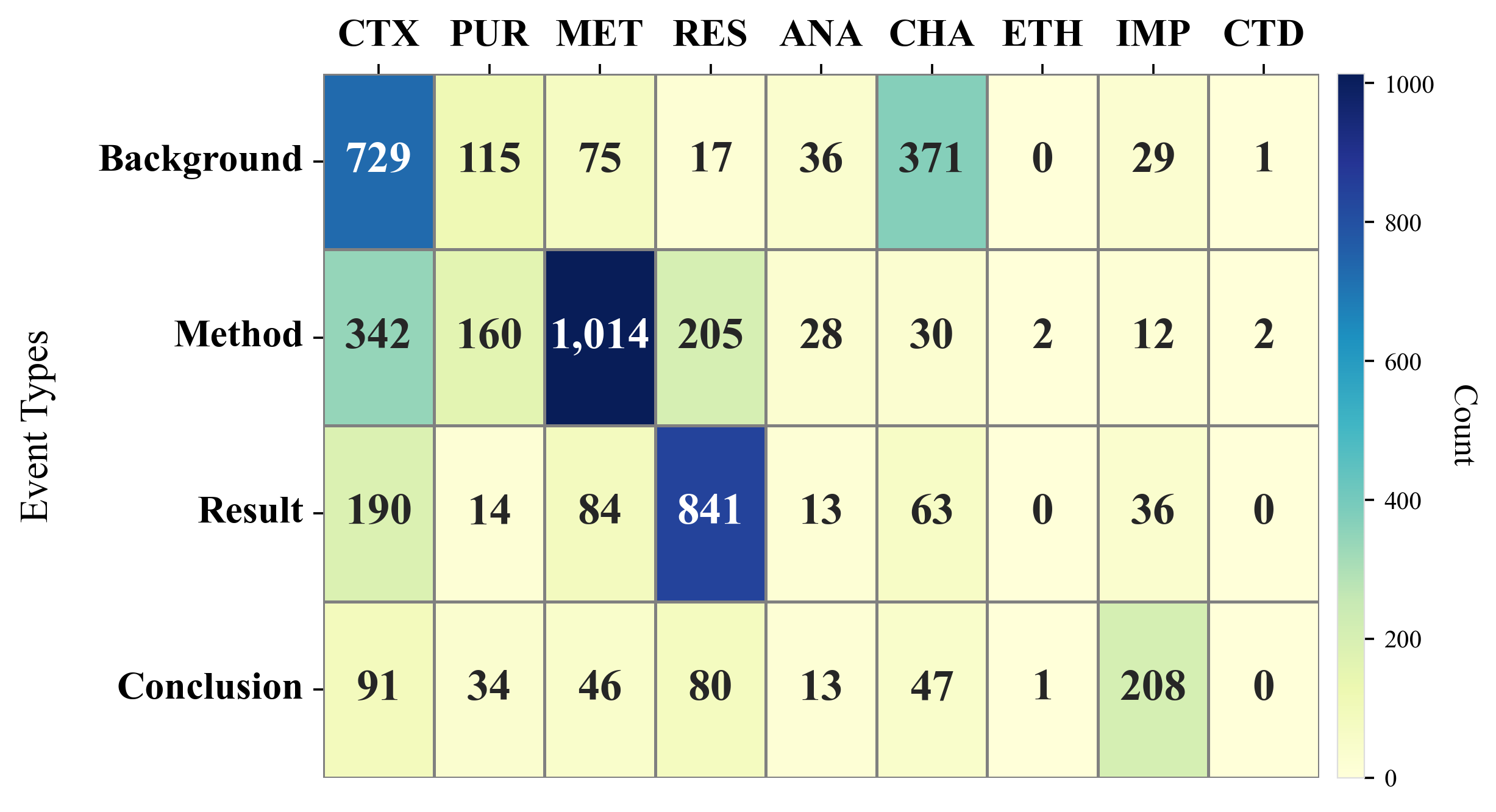}
  \caption{Distribution of argument across event types}
  \label{fig:heatmap2}
  \vspace{-5mm}
\end{figure}

As shown in Figure~\ref{fig:piechart2}, the most frequently annotated arguments are Context (CTX), Method (MET), and Result (RES), highlighting the dataset’s emphasis on core components of scientific reporting. Rare arguments such as Contradictions (CTD) and Ethical (ETH) suggest that such aspects are rarely discussed in the abstracts. The most common event type is the Method, consistent with typical abstracts structures. Moreover, Figure~\ref{fig:heatmap2} shows that argument types align well with event types---for example, Context appears predominantly in Background events, supporting the reliability and internal consistency of our annotations.

\section{Task Definition}
\label{subsec: Task Definition}

Given a document represented as a sequence of sentences $D = \{s_1, \ldots, s_N\}$, our goal is to extract a set of scientific events $E = \{e_1, \ldots, e_M\}$, where each event $e_i$ is a tuple defined as: $e_i = \langle s_{ij}, type_i, \mathrm{Trigger}_i, \mathrm{Arg}_i \rangle$, where $s_{ij}$ denotes a contiguous sentence span in $D$, $type_i$ is the event type, $\mathrm{Trigger}_i$ is the event trigger and $\mathrm{Arg}_i$ consists of event arguments involved in $e_i$. Specifically, we define $\mathrm{Trigger}_i$ as an agent-action-object tuple: 
\vspace{-0.6em}
\[
\mathrm{Trigger}_i = \langle \sigma_{\mathrm{agent}},\ \sigma_{\mathrm{action}},\ \sigma_{\mathrm{object}} \rangle,
\vspace{-0.2em}
\]
where $\sigma \in D$ is a token span that specifies who-does-what in $e_i$, respectively. We further define $\mathrm{Arg}_i$ as a list of argument-role pairs:
\vspace{-0.6em}
\[
\mathrm{Arg}_i = \{a_{ij}, r\},
\vspace{-0.3em}
\]
where $a_{ij} \in D$ denotes the token span that refers to a participating argument entity, and $r$ is the specific role that the argument $a_{ij}$ plays in $e_i$.

To achieve our goal on SciEvent, we define three tasks: (1) event segmentation, (2) trigger identification, and (3) argument extraction. The details of each task are described below.

\paragraph{Task 1: Event Segmentation}
\label{Task 1} This task aims to segment 
any given document $D$ into contiguous sentence spans $\{s_{ij}\}$, with each span $s_{ij}$ corresponding to an event classified under one of four scientific event types $type_i$.

We evaluate model predictions using \textit{Exact Match (EM)} and \textit{Intersection over Union (IoU)} metrics, adapted from span-based evaluation metrics in SemEval \citep{segura-bedmar-etal-2013-semeval} and MUC-5 \citep{chinchor-sundheim-1993-muc}. For each predicted event segment $(\hat{s}_{ij}, \hat{type}_i)$, the above metrics are defined as follow:
\begin{itemize}[topsep=0.5em, itemsep=0.5em, parsep=0em, leftmargin=1.5em]
    \item \textbf{Exact Matching (EM):} $\hat{s}_{ij} = s_{ij}$ and $\hat{type}_i = type_i$
    \item \textbf{Intersection over Union (IoU):} $\frac{|\hat{s_{ij}} \cap s_{ij}|}{|\hat{s_{ij}} \cup s_{ij}|} > 0.5$ and $\hat{type}_i = type_i$.
\end{itemize}
For both strategies, we report Precision (P), Recall (R), and F1-score (F1) over the set of predicted and gold event segments.

\paragraph{Task 2: Trigger Identification}
\label{Task 2}
This task focuses on extracting the trigger for each detected event. As this is a document-level task and scientific events often include multiple candidate triggers, we handle this step separately to explicitly evaluate the model’s ability to correctly identify the core semantic components of an event once its span and type have been identified.

For evaluation, we concatenate each trigger tuple's three components and compute macro ROUGE-L \citep{lin-2004-rouge} between predicted and annotated triggers. Given that ROUGE-L measures the longest common subsequence overlap, we believe that this metric can capture both lexical similarity and structural alignment

\paragraph{Task 3: Argument Extraction}
\label{Task 3}
We decompose this task into two sub-tasks:
\begin{itemize}[topsep=0.3em, itemsep=0.3em, parsep=0em, leftmargin=1.5em]
    \item \textbf{Argument Identification (Arg-I):} Predict the set of argument entity spans $\{a_{ij}\}$ per event.
    \item \textbf{Argument Classification (Arg-C):} Predict the semantic role $r$ associated with each identified argument span $a_{ij}$.
\end{itemize}
We evaluate Arg-I using the F1 score based on the span-matching strategies described in Task \hyperref[Task 1]{1}: \textit{EM} and \textit{IoU} (with a threshold of 0.5). For Arg-C, a prediction is considered correct only if it both matches the gold argument span and also assigns the correct argument role. 

\section{Experiment Settings}
\label{sec:experiment}
\paragraph{Prompting-based LLM baselines} 
We consider four state-of-the-art LLMs as baseline models, including: (1) meta-Llama-3.1-8B-Instruct (Llama) \cite{meta2024llama3}, (2) Qwen2.5-7B-Instruct (Qwen) \cite{qwen2_5}, (3) DeepSeek-R1-Distill-Llama-8B (DS-R1-Llama) \cite{deepseekai2025deepseekr1incentivizingreasoningcapability}, and (4) GPT-4.1 (GPT) \cite{openai2025gpt41}. For Task \hyperref[Task 1]{1}, we conduct a preliminary study on prompt design under the zero-shot manner and used the best prompt adapted from \citealp{sharif-etal-2024-explicit}. For Task \hyperref[Task 2]{2} and \hyperref[Task 3]{3}, we design the prompt template based on a preliminary analysis of existing prompting strategies, including metacognitive prompting \cite{wang-zhao-2024-metacognitive}, instruction-based prompting \cite{sharif-etal-2024-explicit} and paraphrasing these prompts. Considering the risk of model performance being sensitive to the number of examples in the prompt, we test prompts with 0 to 5 examples. We finally employ the prompt adapted from \citealp{sharif-etal-2024-explicit}, which shows consistently better performance across multiple trials. Notably, the prompt content (see details in Appendix~\ref{app:C Prompts}) is derived from our annotation codebook, to ensure a fair comparison between LLMs and humans. 

\vspace{-1mm}
\paragraph{Tuning-based baselines} In addition to prompting-based baselines, we follow prior studies \cite{huang-etal-2024-textee, tong-etal-2022-docee} and also employ turning-based models on our event extraction tasks, adopting three state-of-the-art approaches: (1) DEGREE \citep{hsu-etal-2022-degree}, a data-efficient generative approach to event argument extraction that leverages prompt-based learning for better generalization. (2) OneIE \citep{lin-etal-2020-joint}, a joint information extraction framework that simultaneously performs entity, relation, and event extraction using a unified representation. (3) EE\_QA \citep{du-cardie-2020-event}, a transformer-based model that frames information extraction as a question-answering task, enabling contextualized argument extraction. For all three models, we follow the splitting practice used in prior work \cite{huang-etal-2024-textee} and adopt the same approach and split the training, development, and test sets by document with a ratio of 80\%, 10\%, and 10\%.
\section{Experiment Results}
\label{sec:results}

\paragraph{Scientific event segmentation} 
\begin{table}[H]
\centering
\small
\setlength{\tabcolsep}{3pt}
\begin{tabular}{@{}lcccccc@{}}
% {@{}lccccccl@{}} {|l|ccc|ccc|}
\hline
\multirow{2}{*}{\textbf{Model}} & \multicolumn{3}{c}{\textbf{EM}} & \multicolumn{3}{c}{\textbf{IoU}} \\
\cline{2-7}
& \textbf{P} & \textbf{R} & \textbf{F1} & \textbf{P} & \textbf{R} & \textbf{F1} \\
\hline
DS-R1-Llama & 31.26 & 34.13 & 32.63 & 58.97 & 64.38 & 61.56 \\
Qwen & 43.51 & 36.30 & 39.58 & 70.30 & 58.65 & 63.95 \\
Llama & 38.67 & 31.70 & 34.84 & 62.04 & 50.85 & 55.89 \\
GPT & \textbf{59.07} & \textbf{62.96} & \textbf{60.95} & \textbf{82.98} & \textbf{88.45} & \textbf{85.63} \\
\hline
\end{tabular}
\caption{Scientific event segmentation performance (\%) on zero-shot LLMs using Exact Match (EM) and Intersection over Union (IoU) metrics, showing Precision (P), Recall (R), and F1-score}
\label{tab:chunk_parsing}
\vspace{-4mm}
\end{table}
Table~\ref{tab:chunk_parsing} shows the results of LLM performance under zero-shot manner. We observe that GPT clearly outperforms all others by a wide margin, achieving 60.95\% F1 under EM and 85.63\% under IoU, indicating its strong ability to identify and segment coherent scientific spans. Qwen ranks second, while Llama and DS-R1-Llama trail closely with modest differences. These results suggest that segmentation is best handled by higher-capacity models like GPT.
\vspace{-2mm}
\paragraph{Trigger Identification}
\begin{table}[h]
\centering
\small
\renewcommand{\arraystretch}{1.2}
\setlength{\tabcolsep}{2pt}
\begin{tabular}{@{}lccc@{}}
% {|l|r|r|r|}
\hline
\textbf{Methods} & \textbf{P} & \textbf{R} & \textbf{F1} \\
% \hline
% Human & 81.77 & 84.06 & 81.36 \\
\hline
\textit{Tuning-based models}\\
EEQA & \textbf{81.93} & 34.57 & 45.05 \\
DEGREE & 64.56 & 63.49 & 56.85 \\
OneIE & 73.73 & \textbf{79.40} & 72.40 \\
\hline
\textit{Zero-shot LLMs}\\
DS-R1-Llama & 29.12 & 27.10 & 26.74 \\
Qwen & 43.84 & 55.25 & 47.57 \\
Llama & 54.88 & 61.07 & 55.83 \\
GPT & 65.38 & 72.73 & 67.57 \\
\hline
\textit{One-shot LLMs}\\
DS-R1-Llama & 41.81 & 41.94 & 40.72 \\
Qwen & 56.17 & 68.48 & 59.98 \\
Llama & 53.08 & 63.83 & 56.45 \\
GPT & 72.67 & 77.77 & 74.05 \\
\hline
\textit{Two-shot LLMs}\\
DS-R1-Llama & 34.59 & 36.21 & 34.29 \\
Qwen & 57.27 & 69.71 & 61.18 \\
Llama & 58.94 & 61.18 & 58.34 \\
GPT & 73.38 & 78.45 & 74.76 \\
\hline
\textit{Five-shot LLMs}\\
DS-R1-Llama & 38.63 & 35.63 & 35.18 \\
Qwen & 57.43 & 66.88 & 60.18 \\
Llama & 32.05 & 32.83 & 31.37 \\
GPT & 73.70 & 78.82 & \textbf{75.08} \\
\hline
\end{tabular}
\caption{ROUGE-L scores (\%) for baseline models on the SciEvent trigger identification task, showing Precision (P), Recall (R), and F1.}
\label{tab:ROUGE-L Overall}
\vspace{-3mm}
\end{table}
Table~\ref{tab:ROUGE-L Overall} displays the models’ performance on trigger identification. GPT (five-shot) achieves the best result (F1: 75.08\%), while OneIE also performs competitively (F1: 72.40\%). EEQA exhibits extremely high precision (P: 81.93\%) but poor recall (R: 34.57\%), suggesting over-conservative predictions. Across all LLMs, one-shot prompting consistently improves performance, with DS-R1-Llama showing the largest gain (F1: +13.98\%). While the improvement from zero-shot to one-shot is substantial, surprisingly, further adding more examples yields at most a 1\% gain and can sometimes even reduce performance, particularly for Llama and DS-R1-Llama. Our observation aligns with prior findings that in-context learning may amplify reliance on superficial patterns in demonstrations \cite{min-etal-2022-rethinking}, which is insufficient for the fine-grained comprehension required by event extraction. Accordingly, our subsequent analysis focuses on zero-shot and the overall best-performing one-shot prompts.

\paragraph{Argument Extraction}
\label{section: SciEvent Argument Extraction}
\begin{table}[t]
\centering
\small
\renewcommand{\arraystretch}{1.2}
\setlength{\tabcolsep}{1.5pt}
\begin{tabular}{@{}lcccccc@{}}
% {|l|ccc|ccc|} {@{}lccccccl@{}}
\hline
\multirow{2}{*}{\textbf{Methods}} & \multicolumn{3}{c}{\textbf{Arg-I (IoU)}} & \multicolumn{3}{c}{\textbf{Arg-C (IoU)}} \\
\cline{2-7}
& \textbf{P} & \textbf{R} & \textbf{F1} & \textbf{P} & \textbf{R} & \textbf{F1} \\
% \hline
% Human & 67.52 & 52.46 & 59.04 & 52.33 & 40.65 & 45.76 \\
\hline
\textit{Tuning-based models}\\
EEQA & 32.09 & 33.77 & 32.91 & 25.85 & 27.20 & 26.51 \\
DEGREE & \textbf{67.79} & 19.13 & 29.84 & \textbf{48.99} & 13.83 & 21.57 \\
OneIE & 51.11 & \textbf{56.29} & \textbf{53.57} & 39.69 & \textbf{43.71} & \textbf{41.61} \\
\hline
\textit{Zero-shot LLMs}\\
DS-R1-Llama & 31.11 & 16.46 & 21.53 & 16.32 & 8.63 & 11.29 \\
Qwen & 35.68 & 26.41 & 30.35 & 17.58 & 13.01 & 14.96 \\
Llama & 24.37 & 24.90 & 24.63 & 11.68 & 11.93 & 11.80 \\
GPT & 43.03 & 55.56 & 48.50 & 30.40 & 39.25 & 34.26 \\
\hline
\textit{One-shot LLMs}\\
DS-R1-Llama & 42.62 & 17.67 & 24.98 & 19.59 & 8.12 & 11.48 \\
Qwen & 46.33 & 30.36 & 36.69 & 20.96 & 13.74 & 16.60 \\
Llama & 44.70 & 34.08 & 38.68 & 18.93 & 14.44 & 16.38 \\
GPT & 50.14 & 50.22 & 50.18 & 34.60 & 34.66 & 34.63 \\
\hline
\textit{Two-shot LLMs}\\
DS-R1-Llama & 42.66 & 20.01 & 27.24 & 14.37 & 6.74 & 9.18 \\
Qwen & 46.16 & 31.43 & 37.39 & 21.08 & 14.35 & 17.08 \\
Llama & 40.87 & 25.11 & 31.11 & 18.10 & 11.12 & 13.78 \\
GPT & 49.12 & 51.29 & 50.18 & 33.99 & 35.49 & 34.72 \\
\hline
\textit{Five-shot LLMs}\\
DS-R1-Llama & 36.31 & 20.92 & 26.55 & 13.62 & 7.85 & 9.96 \\
Qwen & 46.94 & 31.36 & 37.60 & 21.67 & 14.48 & 17.36 \\
Llama & 38.36 & 8.93 & 14.49 & 14.98 & 3.49 & 5.66 \\
GPT & 50.04 & 49.93 & 49.98 & 34.51 & 34.42 & 34.47 \\
\hline
\end{tabular}
\caption{IoU-based Precision (P), Recall (R), and F1-score (\%) on baseline models for argument identification (Arg-I) and classification (Arg-C) tasks.}
\label{tab:IoU_Results_Overall}
\vspace{-5mm}
\end{table} Table~\ref{tab:IoU_Results_Overall} reports the performance of all baselines on argument extraction in SciEvent. OneIE achieves the highest scores (Arg-I: 53.57\%, Arg-C: 41.61\%), benefiting from its global features and constraints. DEGREE shows high precision but low recall, indicating that it often misses relevant arguments in scientific abstracts. Among LLMs, GPT (two-shot) performs best (Arg-I: 50.18\%, Arg-C: 34.72\%), while other models perform notably worse, especially on argument classification (Arg-C around 15\%). One-shot prompting provides a modest gain over zero-shot settings, whereas adding more in-context examples shows similar diminishing returns observed in trigger identification. This indicates that merely increasing the number of few-shot examples is insufficient to overcome the fine-grained challenges of scientific argument extraction.
\vspace{-2mm}
\paragraph{Human performance}
\begin{figure}[h]
    \centering
    \includegraphics[width=0.98\linewidth]{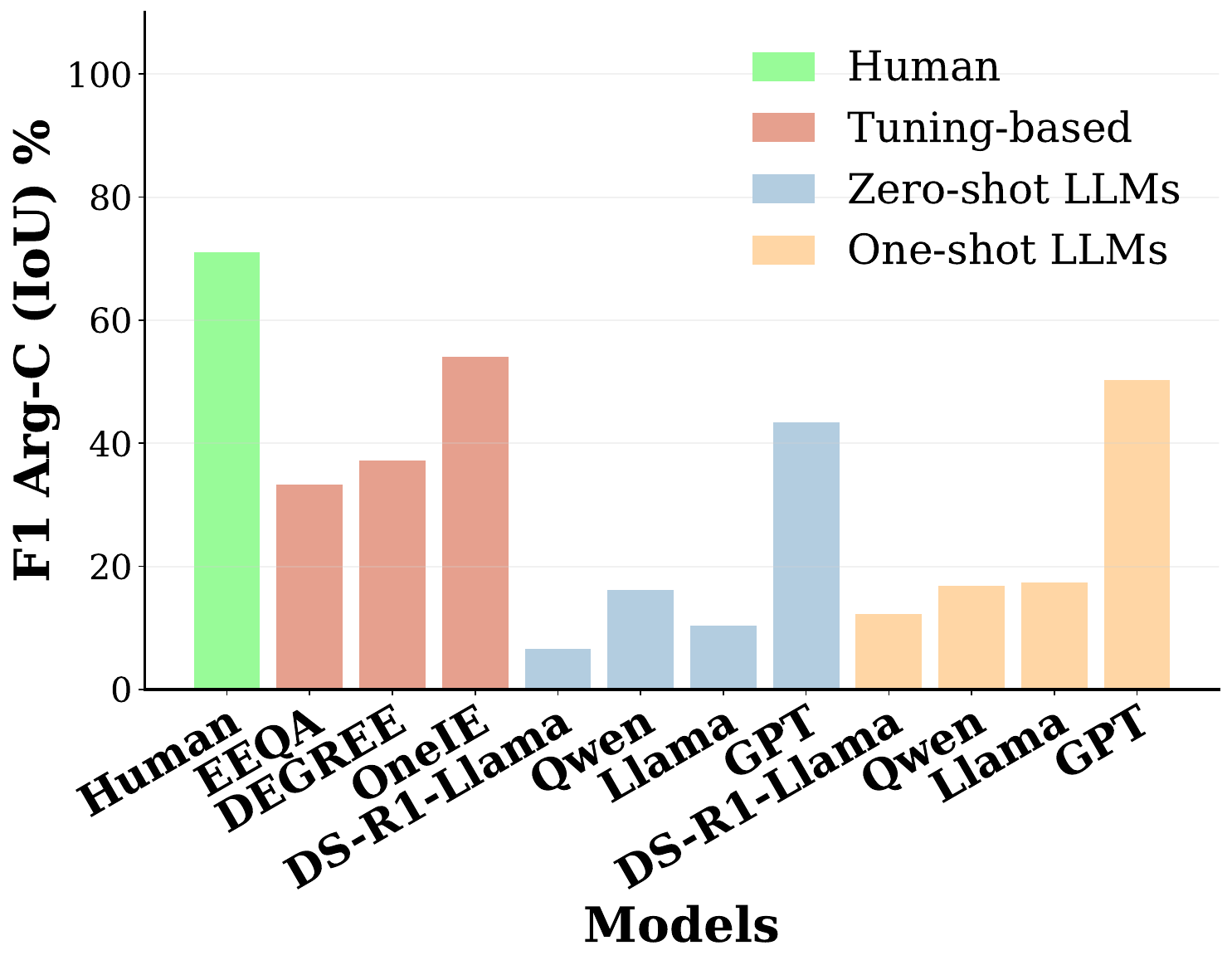}
    \caption{Human performance compared to all baselines on argument classification (Arg-C) using IoU F1 scores.}
    \label{fig:arg_c_performance}
    \vspace{-3mm}
\end{figure}
We compare model and human performance on argument classification. We do not report results for event segmentation, as the Cohen’s kappa score of 0.83 (exact match) on a subset indicates consistently high agreement among annotators, suggesting that event segmentation is relatively unambiguous for humans. As shown in Figure~\ref{fig:arg_c_performance}, there is a substantial gap between human performance and the best model (20\%). This highlights the challenge of multi-domain scientific event extraction and the value of SciEvent for advancing argument level scientific event extraction.
\begin{figure*}[t]
    \centering
    \includegraphics[height=0.25\textheight, width=0.98\linewidth]{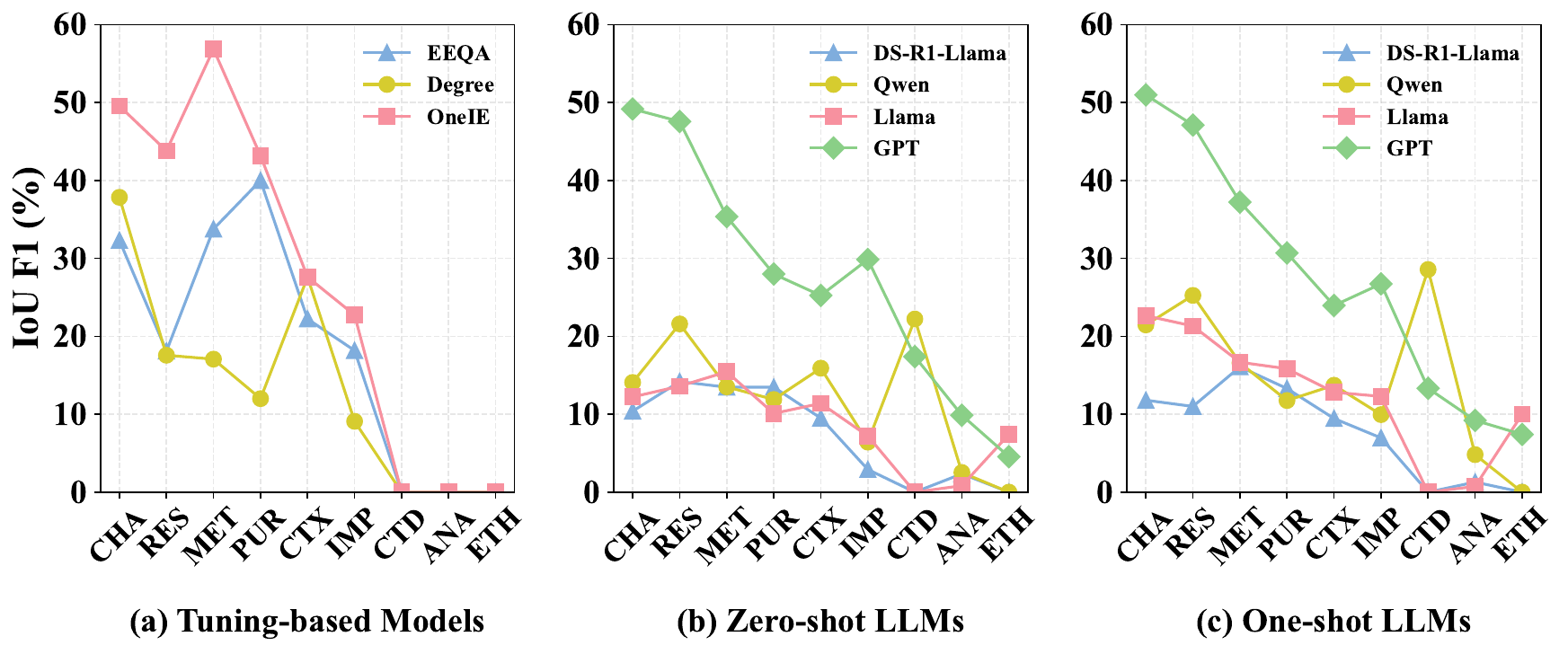}
    \caption{Intersection-over-Union (IoU) on Arg-C F1-scores (\%) across different argument roles for various models on Analysis (ANA), Challenge (CHA), Context (CTX), Method (MET), Purpose (PUR), Result (RES), Ethical (ETH), Implication (IMP), Contradictions (CTD).}
    \label{fig:argument_wise_iou_comparison}
    \vspace{-3mm}
\end{figure*}
\vspace{-4mm}
\paragraph{What is the impact of argument type on argument classification?}
Figure~\ref{fig:argument_wise_iou_comparison} displays IoU-based F1 scores for argument classification across argument roles. Among tuning-based and LLM-based models, OneIE and GPT achieve the strongest performance across nearly all argument roles. Qwen achieves a spike on Contradiction, due to a few correct extractions, but shows worse performance overall. Across all models, Challenge, Result, and Method yield the highest F1 scores, due to their clearer lexical cues and more regular positioning in scientific abstracts. In contrast, arguments like Ethical, Contradiction, and Analysis remain challenging due to data sparsity and a lack of consistent lexical patterns.

\paragraph{What is the impact of event type on argument classification?}
\label{RQ 1}
\begin{figure*}[t]
    \centering
    \includegraphics[height=0.25\textheight, width=0.98\linewidth]{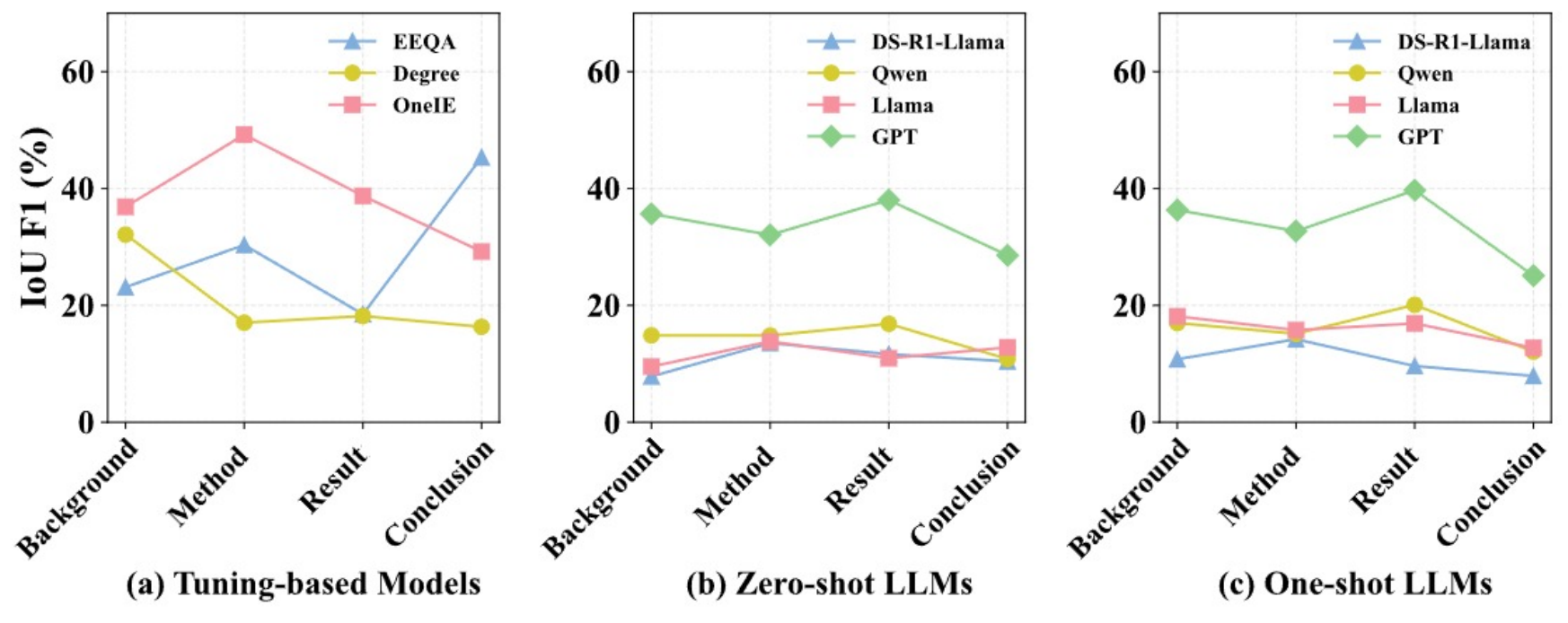}
    \caption{Comparison of Intersection-over-Union (IoU) on Arg-C F1-scores (\%) across different event types for various models on \textit{Background}, \textit{Method}, \textit{Result}, and \textit{Conclusion} events.}
    \label{fig:event_type_wise_iou_comparison}
    \vspace{-3mm}
\end{figure*} 
The arguments in Method exhibit a notable gap: strong performance with supervision (OneIE, EEQA) but poor with zero-/one-shot LLMs on argument classification task (Figure~\ref{fig:event_type_wise_iou_comparison}). This finding suggests that arguments in the Method events are most demanding, due to event's complex structure, arguments' varied phrasing, and dependence on technical details, making performance poorer without supervision. Furthermore, Conclusion shows the lowest Arg-C performance for most models. EEQA performs better because its QA-based templates help extract the implicit and interpretive content typical of Conclusion events. To examine whether event type awareness can improve argument extraction for LLMs, we experiment with incorporating event type information into the prompts. Specifically, we explore two strategies: (1) providing the true event type directly in the prompt, and (2) asking the LLM to first predict the event type and then proceed with argument extraction. Both strategies outperform the original prompt, which lacks event type information, by approximately 2 to 4\% (Table~\ref{tab:true-pred_event_type_iou_comparison}), suggesting that event type awareness enhances LLMs’ performance on our benchmark.

\begin{table}[t]
\centering
\small
\renewcommand{\arraystretch}{1.2}
\setlength{\tabcolsep}{1.5pt}
\begin{tabular}{@{}lcccccc@{}}
\hline
\multirow{2}{*}{\textbf{Methods}} & \multicolumn{3}{c}{\textbf{Arg-I (IoU)}} & \multicolumn{3}{c}{\textbf{Arg-C (IoU)}} \\
\cline{2-7}
& \textbf{P} & \textbf{R} & \textbf{F1} & \textbf{P} & \textbf{R} & \textbf{F1} \\
\hline
\textit{True Event-Type LLMs}\\
DS-R1-Llama & 27.25 & 17.95 & 21.64 & 16.50 & 10.87 & 13.10 \\
Qwen & 31.92 & 35.81 & 33.75 & 16.74 & 18.78 & 17.70 \\
Llama & 17.51 & 27.90 & 21.51 & 9.41 & 14.99 & 11.56 \\
GPT & 42.12 & \textbf{57.13} & 48.49 & 31.29 & \textbf{42.44} & 36.02 \\
\hline
\textit{Pred Event-Type LLMs}\\
DS-R1-Llama & 28.29 & 17.61 & 21.70 & 17.12 & 10.65 & 13.13 \\
Qwen & 31.93 & 35.28 & 33.52 & 17.17 & 18.97 & 18.02 \\
Llama & 20.53 & 28.71 & 23.94 & 10.36 & 14.48 & 12.08 \\
GPT & \textbf{44.42} & 55.22 & \textbf{49.23} & \textbf{32.51} & 40.42 & \textbf{36.04} \\
\hline
\end{tabular}
\caption{IoU-based Precision (P), Recall (R), and F1-score (\%) comparing argument identification (Arg-I) and classification (Arg-C) performance given true event type or predicted event type.}
\label{tab:true-pred_event_type_iou_comparison}
\vspace{-5mm}
\end{table}

\vspace{-3mm}
\paragraph{What is the impact of scientific domains on argument classification?}
\label{RQ 2}
\begin{figure*}[h]
    \centering
    \includegraphics[height=0.23\textheight, width=0.98\linewidth]{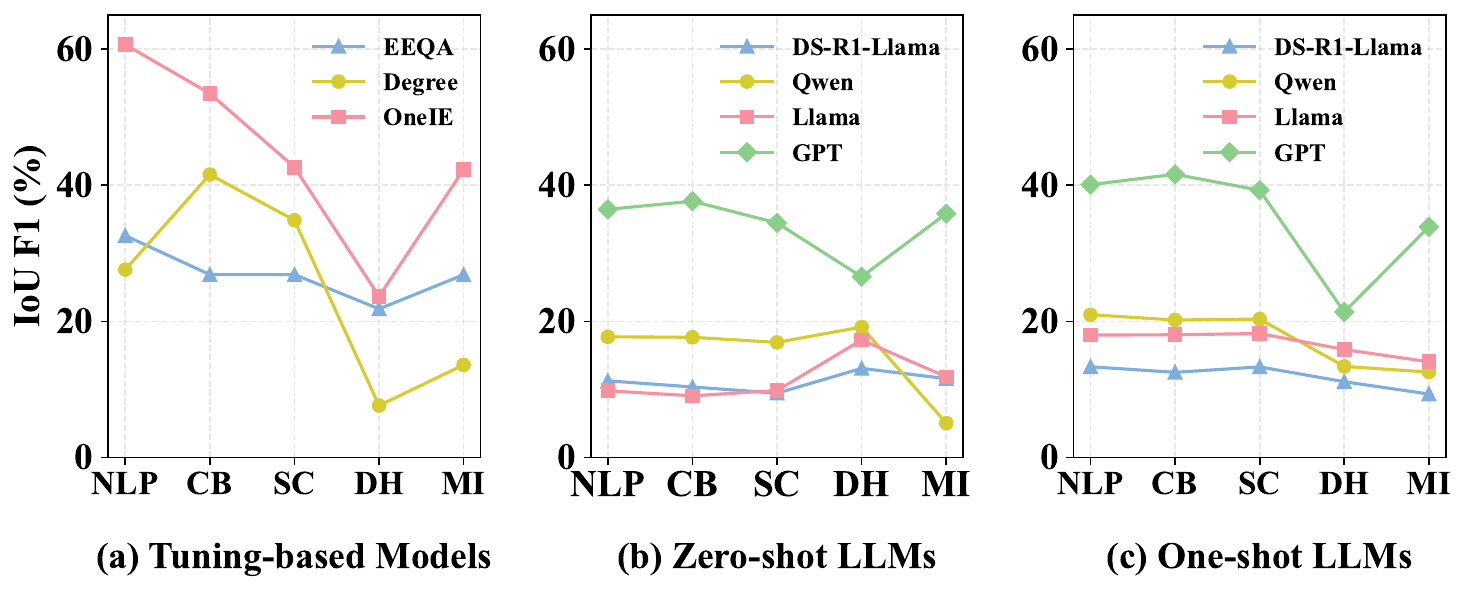}
    \caption{Comparison of Intersection-over-Union (IoU) on Arg-C F1-scores (\%) across different academic domains for various models on Natural Language Processing (NLP), Computational Biology (CB), Social Computing (SC), Digital Humanities (DH), and Medical Informatics (MI).}
    \label{fig:venue_wise_iou_comparison}
    \vspace{-3mm}
\end{figure*}
In the argument classification task (Figure~\ref{fig:venue_wise_iou_comparison}), Natural Language Processing and Computational Biology domains yield the highest F1 scores, benefiting from consistent linguistic patterns and clearer argument structures. In contrast, Digital Humanities and Medical Informatics present greater challenges, due to varied rhetorical styles and longer, denser abstracts, respectively.

\vspace{-3mm}
\paragraph{How does removal of domain affect performance?}
\label{RQ 3}
\begin{figure}[h!]
    \centering
    \includegraphics[width=0.98\linewidth]{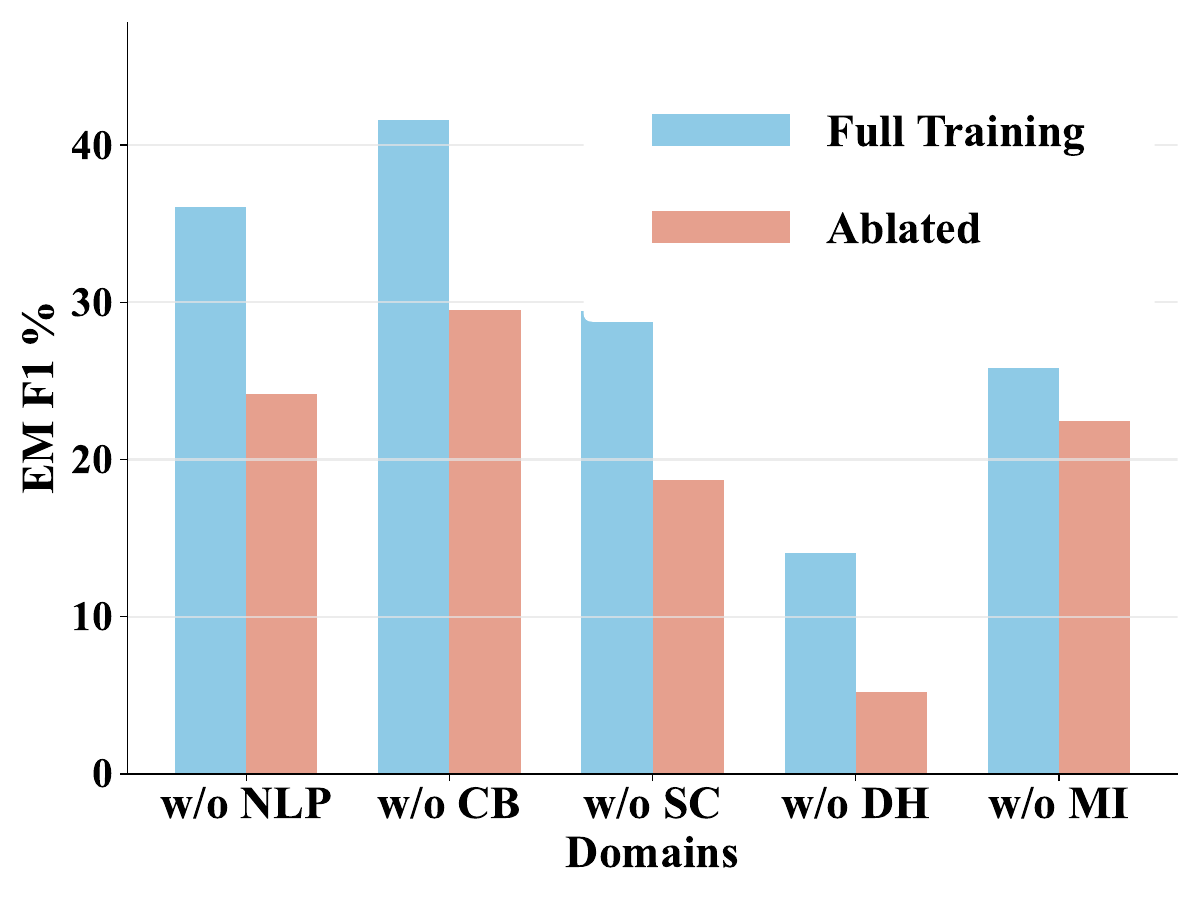}
    \caption{Arg-C F1-scores reported for full training versus training with one domain removed for OneIE under Exact Match (EM).}
    \label{fig:oneie_em_performance}
    \vspace{-3mm}
\end{figure}
We compare the argument classification performance of the OneIE model under the Exact Match (EM) setting using the full training set versus ablated training sets (Figure \ref{fig:oneie_em_performance}). Removing a domain from training data leads to a noticeable drop in its corresponding performance, confirming that domain-specific knowledge contributes directly to accurate argument classification. The largest declines are observed in Digital Humanities and Computational Biology, indicating that these domains contain more unique or specialized linguistic patterns that are not easily generalized from other domains. In contrast, Medical Informatics shows relatively smaller drop, suggesting better generalizability or partial overlap with language patterns present in the other domains.
\section{Conclusion}
\label{sec:conclusion}
In this paper, we introduce SciEvent, a novel benchmark for SciIE across multiple domains. By framing scientific texts as a sequence of universal events and corresponding fine-grained arguments, SciEvent provides a unified and domain-independent structure for representing scientific information. Specifically, We develop an annotation pipeline comprising event segmentation and trigger-argument extraction, and defined three corresponding tasks: (1) event segmentation, (2) trigger identification, and (3) argument extraction. Our benchmark covers five diverse domains with manual annotations, enabling robust evaluation of Event Extraction. Experiments on diverse state-of-the-art tuning-based Event Extraction systems and tuning-free LLMs show clear performance gaps ($\sim$20\%) between model predictions and human annotations, especially on argument classification task. SciEvent supports applications such as knowledge graph construction, cross-domain literature review, and scientific summarization. It provides a challenging testbed for extracting nuanced scientific information, benefiting both NLP researchers for advancing event extraction methodology and evaluating cross-domain generalization, and interdisciplinary scholars for accelerating literature review, synthesizing findings, and generating domain knowledge resources.

\newpage
\section*{Limitations}
One limitation of our work is the potential for data contamination in large language models, as our dataset is constructed from recent publications (mostly from 2023, and 2021 to 2023 for Digital Humanities), which may overlap with LLM pretraining corpora. Nevertheless, we emphasize that our benchmark offers a novel formulation by representing scientific abstracts as structured sequences of events, enabling a context-aware capture of key scientific information. This event-centric SciIE schema is novel, and current LLMs lack training to extract scientific content in this structured manner. Additionally, SciEvent is built on abstracts only, which, while concise and widely available, may omit key discourse elements found in full papers limiting applicability to document-level information extraction. In future work, we plan to extend SciEvent to include full papers to better support comprehensive scientific IE, and also consider more event types and arguments roles since the full paper can contain more information such as Assumptions.
\section*{Ethical Considerations}
We provide details about compensation rate for annotators. We recruited eleven graduate students in total and provided a compensation rate of \$12.80 per hour. This rate applied to both gold-standard annotation and human performance baseline annotations. 

\section*{Acknowledgement} 
We thank the anonymous reviewers for their insightful comments and helpful suggestions. We are also grateful to all annotators for their contributions to this work. Finally, we acknowledge the support of the Institute of Integrative Artificial Intelligence at Indiana University.

% % 1. Statistics
% % 2. annotation method (source, scheme, tool)
% % 3. Quality of data (Agreement, pretest, speed, cost)
% % 4. Comparison to existing datasets (scierc, scier, scirex)
% % 5. 

% \section{Experiment} 
%TODO, find citation here. 
%\section{Experiment}

% \subsection{Evaluation Settings}

% We evaluate on four tasks: trigger identification, trigger classification, argument identification and argument classification. 

% For all four tasks, we use K shot in context learning setting for generation agents (LLaMA, GPT, ChatGPT), and N\% of all training data for supervised models (BERT\_QA, OneIE and DEGREE), where $K\in \{0,1,3,5\}$ and $N\in \{10, 20, 50, 100\}$
% Trigger identification, Trigger classification,
% Argument identification, Argument classificiation
% 0,1,3,5 shots LLMs
% 10%, 30%, 50% data for supervised models
%\section{Results and Analysis}

%for all baselines %read more on domain transfer paper
%read the evaluation papers to see what aspects they are evaluating on, and then design ablation study.

% What to remove???
% Remove certain domain, and see how it affects the model's performance on that domain. Remove 1 domain, remove 4 domain.
% Remove Key arguments (agent+trigger+object)
% Remove additional arguments (context, method, etc.)

% \subsection{Error Analysis}

\bibliography{anthology}
\bibliographystyle{acl_natbib}

\appendix
\label{sec:appendix}

\section{Domain-wise Arguments Distribution Analysis}
\begin{figure}[h]
  \centering
  \includegraphics[width=0.98\linewidth]{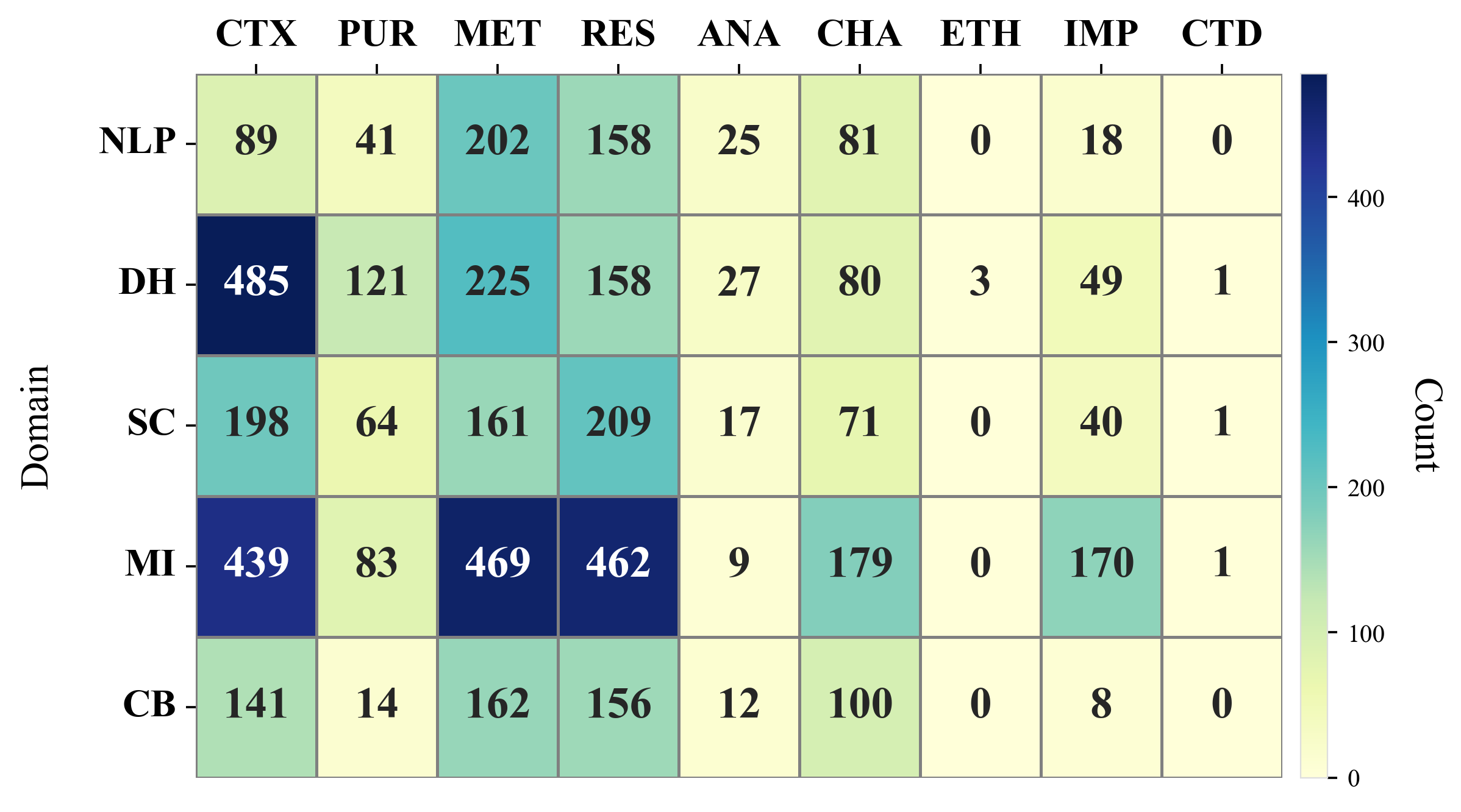}
  \caption{Distribution of argument types across all domains}
  \label{fig:heatmap1}
  \vspace{-5mm}
\end{figure}
We report the distribution of argument types across scientific domains (Figure~\ref{fig:heatmap1}). While all domains emphasize Results, Digital Humanities (DH) is a notable exception, being dominated by Context arguments. Among STEM domains---Natural Language Processing (NLP), Computational Biology (CB), and Medical Informatics (MI)---Method arguments are the most prevalent, reflecting their methodological focus. In contrast, DH and Social Computing (SC) place more emphasis on Context and Results, respectively, aligning with the rhetorical nature of these fields. Notably, MI contains the highest number of arguments overall, likely due to the length of its abstracts, even though fewer were annotated to balance domain coverage.

\section{Codebook details}
\label{app:codebook}
\subsection{Annotation Tool}
\label{app:tool}
\begin{figure*}[h]
  \centering
  \includegraphics[width=0.98\linewidth]{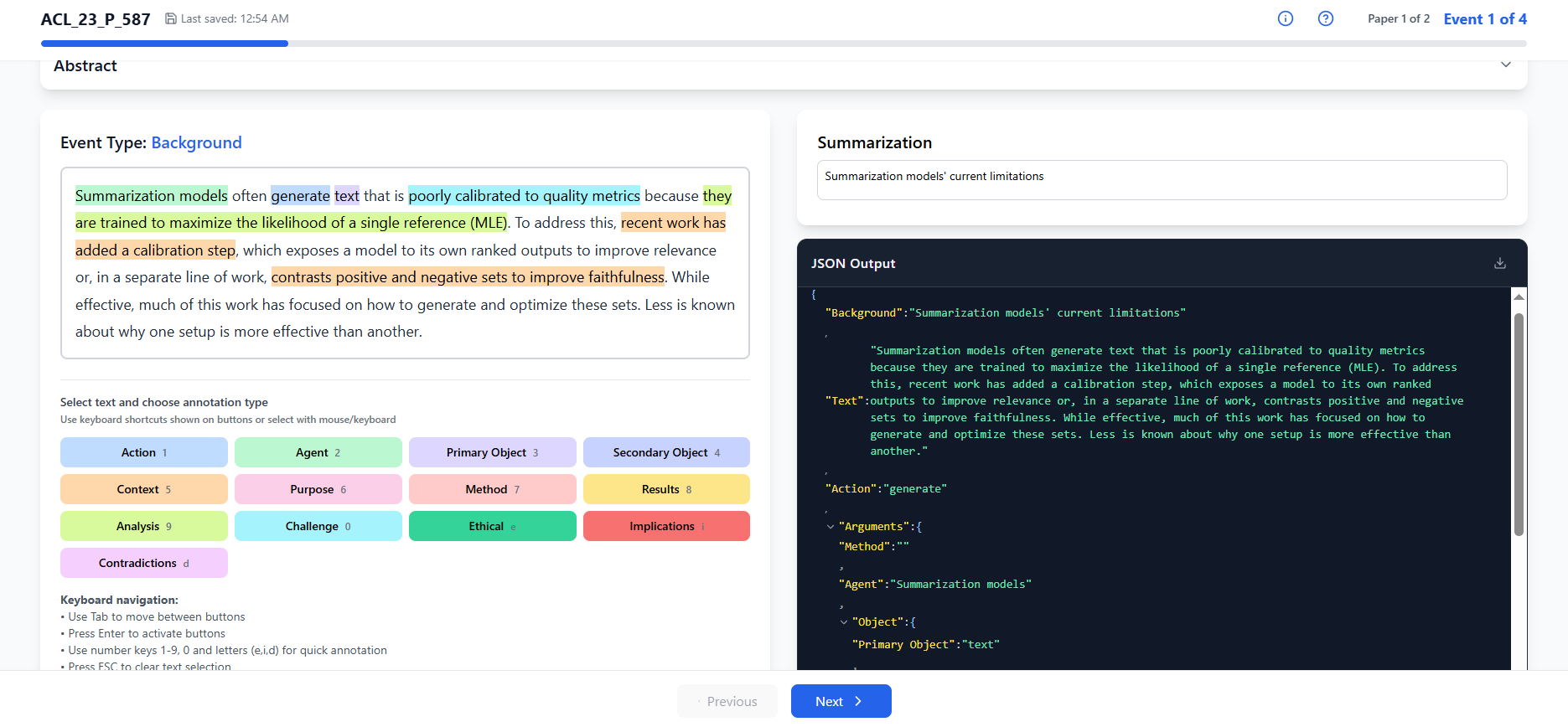}
  \caption{Annotation Tool Interface}
  \label{fig:tool interface}
  \vspace{-5mm}
\end{figure*}

We deploy our annotation tool on Render\footnote{\url{https://render.com/}}. Figure~\ref{fig:tool interface} shows our annotation interface.
\subsection{Event Type Definition}
\begin{itemize}
  \item \textbf{Background:} Briefly outlines the context, motivation, and problem being addressed. It highlights the research gap and the paper's objectives or research questions.
  
  \item \textbf{Method:} Summarizes the methodologies, frameworks, or techniques used to conduct the study, including experimental setups, algorithms, datasets, or analytical tools.
  
  \item \textbf{Result:} Reports the main outcomes of the research, emphasizing key data, trends, or discoveries. Focuses on what was achieved or learned.
  
  \item \textbf{Conclusion:} Discusses the significance of the findings, their impact on the field, potential applications, and how they address the initial problem or research gap. May include recommendations or future research directions.
\end{itemize}

\subsubsection{Trigger Definition}
\begin{itemize}
  \item \textbf{Action:} The most representative verb or verb phrase in the event, including auxiliary verbs like \textit{am}, \textit{is}, \textit{are}, \textit{have}, and \textit{has}.
  
  \item \textbf{Agent:} The entity responsible for initiating or performing the Action, such as a person, system, method, or organization.
  
  \item \textbf{Object:} The entity that receives, is affected by, or is the focus of the Action (e.g., a concept, result, or entity being acted upon). During annotation, Objects may be separated; in such cases, annotate them as \textit{Primary Object} and \textit{Secondary Object}. Include only the Object spans themselves, and do not include the separators or intervening material.
\end{itemize}

\subsection{Argument Definition}
\begin{itemize}
  \item \textbf{Context}\\
  \textbf{Definition:} Provides foundational or situational information of the event.\\
  \textbf{Example:} \textbf{Deep learning} has revolutionized \textbf{natural language processing tasks}, enabling state-of-the-art results in translation, summarization, and question answering.
  \item \textbf{Purpose}\\
  \textbf{Definition:} Defines the purpose or aim of the event.\\
  \textbf{Example:} This study aims to \textbf{develop a lightweight transformer model} suitable for deployment on edge devices.
  \item \textbf{Method}\\
  \textbf{Definition:} Techniques, tools, methodology, or frameworks used in the event.\\
  \textbf{Example:} We employ \textbf{a combination of knowledge distillation and parameter pruning} to reduce model size while maintaining accuracy.
  \item \textbf{Result}\\
  \textbf{Definition:} Observations or outputs of the event.\\
  \textbf{Example:} The proposed method achieves a \textbf{40\% reduction in model size} with only a \textbf{1\% drop in accuracy on the GLUE benchmark}.
  \item \textbf{Analysis}\\
  \textbf{Definition:} Interpretation or explanation of other arguments.\\
  \textbf{Example:} The slight decrease in accuracy can be attributed to \textbf{the removal of redundant parameters} that minimally impact overall model performance.
  \item \textbf{Challenge}\\
  \textbf{Definition:} Constraints or weaknesses of the context, method, or results.\\
  \textbf{Example:} One significant limitation of the approach is its \textbf{dependency on large-scale labeled datasets} for effective knowledge distillation.
  \item \textbf{Ethical}\\
  \textbf{Definition:} Ethical concerns, implications, and justifications of the event.\\
  \textbf{Example:} The deployment of these models must address concerns about potential biases in training data that could \textbf{unfairly disadvantage certain user groups}.
  \item \textbf{Implication}\\
  \textbf{Definition:} Broader applicability, significance, or potential for future research.\\
  \textbf{Example:} Our approach opens the door for \textbf{deploying advanced NLP models} on low-power devices, paving the way for \textbf{accessible AI in remote or resource-constrained environments}.
  \item \textbf{Contradiction}\\
  \textbf{Definition:} Disagreements with existing knowledge.\\
  \textbf{Example:} Contrary to previous studies suggesting that \textbf{parameter pruning significantly reduces accuracy}, our results demonstrate \textbf{minimal performance loss with careful pruning strategies}.
\end{itemize}
\subsection{Additional Annotation Rules}
\label{app: Annotation Rule}
\begin{itemize}
  \item \textbf{Annotate by Breaking Down Sentences:}  
  Please annotate segments of a sentence (a part of a sentence) instead of a full sentence if different segments of the sentence can be fit into different arguments.
  
  \item \textbf{Passive Tense:}  
  In a passive tense structure: Something (Agent) + is done (Passive Verb) + by Someone/Something (Object).
  
  \item \textbf{Indirect Object:}  
  If there is no direct object, you should leave the Object empty.
  
  \item \textbf{Entire Clause as Object:}  
  In the following structure, the entire clause is the \textless Object\textgreater:  
  \textless Agent\textgreater\ + \textless Actions like: show, demonstrate, illustrate, prove, found, explain, indicate, conclude, etc.\textgreater\ + that / what / who / which / where / when / how / whether + clause.
  
  \item \textbf{Text that Fits Multiple Arguments:}  
  If a text span can fit into multiple \textless Arguments\textgreater, follow this order of importance:  
  \textbf{Results} > Purpose > Method > Analysis > Implication > Challenge > Contradiction > Context > Ethical.  
  Results is the most important, and Ethical is the least.
  
  \item \textbf{Abbreviation:}  
  You should use both the original term and its abbreviation when both are given together, e.g., Chain-of-Thought (CoT), not only Chain-of-Thought or CoT.
  
  \item \textbf{Use of Primary and Secondary Object:}  
  Annotate as \textit{Primary Object} and \textit{Secondary Object} when an Object is expressed in two separate spans within a sentence. Two common cases occur:

  (1) \textbf{Parallel Objects:} The structure is \textit{Action + Primary Object + (and / as well as / also) + Secondary Object}. Intervening clauses (e.g., ``which \ldots'', ``that \ldots'') may appear between Objects; ignore these clauses and annotate only the Object spans.  
  \textit{Example:} ``We analyze protein sequences, which exhibit structural variation, and gene expression profiles.''  
  Annotation: <Primary Object>: protein sequences; <Secondary Object>: gene expression profiles.

  (2) \textbf{Transformation Objects:} The structure is \textit{Action + Primary Object + (into / to / for) + Secondary Object}. The \textit{Primary Object} is what is transformed, and the \textit{Secondary Object} is where it is mapped or placed. Always include the preposition introducing the Secondary Object.  
  \textit{Example:} ``We aimed to map the most frequently discussed factors into health systems and practical use.''  
  Annotation: <Primary Object>: the most frequently discussed factors; <Secondary Object>: into health systems and practical use.

  In both cases, please only annotate separate spans of Primary Object and Secondary Object, DO NOT include intervening descriptive clauses or anything in between.
  
\end{itemize}

\clearpage
% Define a single JSON style for all prompts
\lstdefinestyle{jsonprompt}{
    basicstyle=\ttfamily\scriptsize,
    breaklines=true,
    breakatwhitespace=false,
    breakindent=0pt,
    columns=flexible,
    keepspaces=true,
    showstringspaces=false,
    tabsize=2,
    backgroundcolor=\color{white},
    frame=none,
    literate={\ \ }{{\ }}1  % Preserve spaces
}
\onecolumn  
\section{Prompts} 
\label{app:C Prompts}
Considering the sensitivity of LLM performance to prompt phrasing, we explore a variety of prompt variations to identify the optimal one. These variations include incorporating the Metacognitive Prompting technique \cite{wang-zhao-2024-metacognitive}, adopting role definitions with direct information extraction instead of a QA format, and paraphrasing prompt instructions—for example, replacing “\#\#\# Output (JSON only)” with “\#\#\# Your Answer (JSON format).” Each prompt version was evaluated in at least three runs to assess prediction stability, and the final prompt was selected based on its consistent performance across trials.

In this section, we present the prompt designs for each task. We include the Zero-Shot prompt for \textit{Scientific Abstract Segmentation} and \textit{Trigger Identification \& Argument Extraction}, and the One-Shot prompt for \textit{Trigger Identification \& Argument Extraction}. Additionally, we provide the two event type awareness prompt as well, (1) \textit{True Event-Type Trigger Identification \& Argument Extraction}
and (2) \textit{Predict Event-Type Trigger Identification \& Argument Extraction}.
\begin{tcolorbox}[
    enhanced,
    breakable,
    width=\textwidth,
    enlarge left by=0mm,
    enlarge right by=0mm,
    title={Zero-Shot Scientific Abstract Segmentation Prompt},
    colback=white,
    colframe=black,
    fonttitle=\bfseries
]
You are a strict extraction assistant. Never explain, never repeat, only extract in the required format.

\noindent \textbf{\# \# \# Abstract: \# \# \#}

\noindent \{abstract\}

\noindent \textbf{\# \# \# Extraction Rules: \# \# \#}
\begin{itemize}
\item Copy full, continuous sentences from the abstract. No changes, summaries, or guessing allowed.
\item Each sentence must belong to only one section.
\item Sections must use continuous text spans. No skipping around.
\item If no content fits a section, output exactly <NONE>.
\item No explanations, no extra text, no format changes.
\end{itemize}

\noindent \textbf{\# \# \# Section Definitions: \# \# \#}
\begin{itemize}
\item \textbf{Background}: Problem, motivation, context, research gap, or objectives.
\item \textbf{Method}: Techniques, experimental setups, frameworks, datasets.
\item \textbf{Result}: Main findings, discoveries, statistics, or trends.
\item \textbf{Conclusion}: Importance, impact, applications, or future work.
\end{itemize}

\noindent \textbf{\# \# \# Exact Output Format: \# \# \#}

\begin{lstlisting}[style=jsonprompt]
[Background]: <EXACT TEXT or <NONE>>

[Method]: <EXACT TEXT or <NONE>>

[Result]: <EXACT TEXT or <NONE>>

[Conclusion]: <EXACT TEXT or <NONE>>
\end{lstlisting}
\end{tcolorbox}

\clearpage % Force start on a new page

\begin{tcolorbox}[
    enhanced,
    breakable,
    width=\textwidth,
    enlarge left by=0mm,
    enlarge right by=0mm,
    title={Zero-Shot \text{Trigger Identification \& Argument Extraction} Prompt},
    colback=white,
    colframe=black,
    fonttitle=\bfseries
]
You are an expert argument annotator. Given a part of a scientific abstract, you need to identify the key trigger for the event (the main verb or action that signals an important research activity) and annotate the abstract with the corresponding argument components related to this trigger. Extractions should capture complete phrases around this key trigger and be organized in a single JSON format, containing only what is explicitly stated in the text without adding any interpretation.

\noindent \textbf{\# \# \# Abstract Segment to Analyze:}

\noindent \{abstract\}

\noindent \textbf{\# \# \# Argument Components to Extract:}

\noindent \textbf{Action:} What is the SINGLE most representative trigger (verb or verb phrase) in the segment? 

\noindent \textbf{Agent:} Who or what is performing the Action? 

\noindent \textbf{Object:}
\begin{itemize}
\item \textbf{Primary Object:} What is directly receiving or affected by the Action? 
\item \textbf{Secondary Object:} What is a secondary entity also receiving the Action?
\end{itemize}

\noindent \textbf{Context:} What provides foundational or situational information of the event?

\noindent \textbf{Purpose:} What is the purpose or aim of the event?

\noindent \textbf{Method:} What techniques, tools, approaches, or frameworks are used in the event?

\noindent \textbf{Results:} What are the outcomes, observations or findings of the event?

\noindent \textbf{Analysis:} What are the interpretations or explanations of other arguments?

\noindent \textbf{Challenge:} What are the constraints or weaknesses of the event?

\noindent \textbf{Ethical:} What are the ethical concerns, justifications or implications of the event?

\noindent \textbf{Implications:} What is the broader significance or potential for future applications/research?

\noindent \textbf{Contradictions:} What are the disagreements with existing knowledge?

\noindent \textbf{\# \# \# Extraction Rules:}

\begin{enumerate}
\item Extract complete phrases, not just single words.
\item Only extract elements that are explicitly present. Mark missing elements as [\texttt{"<NONE>"}].
\item Use the exact text from the abstract.
\item Break down sentences when different parts fit different arguments.
\item NEVER use the same span of text for multiple arguments - each piece of text must be assigned to exactly one argument type. However, multiple text spans can be part of the same argument (e.g., [\texttt{"text span 1"}, \texttt{"text span 2".....}] can be used for a single argument type) if different parts of the text contribute to the same argument.
\item If text could fit multiple arguments, prioritize in this order: Results > Purpose > Method > Analysis > Implication > Challenge > Contradiction > Context > Ethical
\end{enumerate}

\noindent \textbf{\# \# \# Output Format:}

\begin{lstlisting}[style=jsonprompt]
{
  "Action": "EXACT TEXT or <NONE>",
  "Agent": ["EXACT TEXT or <NONE>"],
  "Object": {
    "Primary Object": ["EXACT TEXT or <NONE>"],
    "Secondary Object": ["EXACT TEXT or <NONE>"]
  },
  "Context": ["EXACT TEXT or <NONE>"],
  "Purpose": ["EXACT TEXT or <NONE>"],
  "Method": ["EXACT TEXT or <NONE>"],
  "Results": ["EXACT TEXT or <NONE>"],
  "Analysis": ["EXACT TEXT or <NONE>"],
  "Challenge": ["EXACT TEXT or <NONE>"],
  "Ethical": ["EXACT TEXT or <NONE>"],
  "Implications": ["EXACT TEXT or <NONE>"],
  "Contradictions": ["EXACT TEXT or <NONE>"]
}
\end{lstlisting}

\noindent \textbf{\# \# \# IMPORTANT INSTRUCTIONS:}
\begin{itemize}
\item You MUST return ONLY ONE JSON structure.
\item NO explanation text, thinking, or commentary before or after the JSON.
\item NEVER repeat the JSON structure.
\item ALL fields must use arrays with [\texttt{"<NONE>"}] for missing arguments.
\item Follow the EXACT format shown in the template.
\item ONLY extract arguments that are explicitly present in the text. DO NOT hallucinate or add any information not found in the abstract.
\end{itemize}

\noindent \textbf{\# \# \# Output (JSON only)}
\end{tcolorbox}

\clearpage % Force a page break after the second prompt

\begin{tcolorbox}[
    enhanced,
    breakable,
    width=\textwidth,
    enlarge left by=0mm,
    enlarge right by=0mm,
    title={One-Shot  \text{Trigger Identification \& Argument Extraction} Prompt},
    colback=white,
    colframe=black,
    fonttitle=\bfseries
]
You are an expert argument annotator. Given a part of a scientific abstract, you need to identify the key trigger for the event (the main verb or action that signals an important research activity) and annotate the abstract with the corresponding argument components related to this trigger. Extractions should capture complete phrases around this key trigger and be organized in a single JSON format, containing only what is explicitly stated in the text without adding any interpretation.

\noindent \textbf{\# \# \# Abstract Segment to Analyze:}

\noindent \{abstract\}

\noindent \textbf{\# \# \# Argument Components to Extract:}

\noindent \textbf{Action:} What is the SINGLE most representative trigger (verb or verb phrase) in the segment? 

\noindent \textbf{Agent:} Who or what is performing the Action? 

\noindent \textbf{Object:}
\begin{itemize}
\item \textbf{Primary Object:} What is directly receiving or affected by the Action? 
\item \textbf{Secondary Object:} What is a secondary entity also receiving the Action?
\end{itemize}

\noindent \textbf{Context:} What provides foundational or situational information of the event?

\noindent \textbf{Purpose:} What is the purpose or aim of the event?

\noindent \textbf{Method:} What techniques, tools, approaches, or frameworks are used in the event?

\noindent \textbf{Results:} What are the outcomes, observations or findings of the event?

\noindent \textbf{Analysis:} What are the interpretations or explanations of other arguments?

\noindent \textbf{Challenge:} What are the constraints or weaknesses of the event?

\noindent \textbf{Ethical:} What are the ethical concerns, justifications or implications of the event?

\noindent \textbf{Implications:} What is the broader significance or potential for future applications/research?

\noindent \textbf{Contradictions:} What are the disagreements with existing knowledge?

\noindent \textbf{\# \# \# Extraction Rules:}

\begin{enumerate}
\item Extract complete phrases, not just single words.
\item Only extract elements that are explicitly present. Mark missing elements as [\texttt{"<NONE>"}].
\item Use the exact text from the abstract.
\item Break down sentences when different parts fit different arguments.
\item NEVER use the same span of text for multiple arguments - each piece of text must be assigned to exactly one argument type. However, multiple text spans can be part of the same argument (e.g., [\texttt{"text span 1"}, \texttt{"text span 2".....}] can be used for a single argument type) if different parts of the text contribute to the same argument.
\item If text could fit multiple arguments, prioritize in this order: Results > Purpose > Method > Analysis > Implication > Challenge > Contradiction > Context > Ethical
\end{enumerate}

\noindent \textbf{Here is a one-shot example of a complete abstract:}

\noindent \textbf{Background Event}\\
For abstract: "Second language acquisition (SLA) research has extensively studied cross-linguistic transfer, the influence of linguistic structure of a speaker's native language [L1] on the successful acquisition of a foreign language [L2]. Effects of such transfer can be positive (facilitating acquisition) or negative (impeding acquisition). We find that NLP literature has not given enough attention to the phenomenon of negative transfer."

\noindent Output:
\begin{lstlisting}[style=jsonprompt]
{
  "Action": "has extensively studied",
  "Agent": ["Second language acquisition (SLA) research"],
  "Object": {
    "Primary Object": ["cross-linguistic transfer"],
    "Secondary Object": ["<NONE>"]
  },
  "Context": ["Effects of such transfer can be positive (facilitating acquisition) or negative (impeding acquisition)"],
  "Purpose": ["<NONE>"],
  "Method": ["<NONE>"],
  "Results": ["<NONE>"],
  "Analysis": ["<NONE>"],
  "Challenge": ["We find that NLP literature has not given enough attention to the phenomenon of negative transfer"],
  "Ethical": ["<NONE>"],
  "Implications": ["<NONE>"],
  "Contradictions": ["<NONE>"]
}
\end{lstlisting}

\noindent \textbf{Method Event}\\
For abstract: "To understand patterns of both positive and negative transfer between L1 and L2, we model sequential second language acquisition in LMs. Further, we build a Multilingual Age Ordered CHILDES (MAO-CHILDES) — a dataset consisting of 5 typologically diverse languages, i.e., German, French, Polish, Indonesian, and Japanese — to understand the degree to which native Child-Directed Speech (CDS) [L1] can help or conflict with English language acquisition [L2]."

\noindent Output:
\begin{lstlisting}[style=jsonprompt]
{
  "Action": "model",
  "Agent": ["we"],
  "Object": {
    "Primary Object": ["sequential second language acquisition in LMs"],
    "Secondary Object": ["<NONE>"]
  },
  "Context": ["<NONE>"],
  "Purpose": ["To understand patterns of both positive and negative transfer between L1 and L2"],
  "Method": ["we build a Multilingual Age Ordered CHILDES (MAO-CHILDES)"],
  "Results": ["<NONE>"],
  "Analysis": ["a dataset consisting of 5 typologically diverse languages, i.e., German, French, Polish, Indonesian, and Japanese"],
  "Challenge": ["<NONE>"],
  "Ethical": ["<NONE>"],
  "Implications": ["<NONE>"],
  "Contradictions": ["<NONE>"]
}
\end{lstlisting}

\noindent \textbf{Result Event}\\
For abstract: "To examine the impact of native CDS, we use the TILT-based cross lingual transfer learning approach established by Papadimitriou and Jurafsky (2020) and find that, as in human SLA, language family distance predicts more negative transfer. Additionally, we find that conversational speech data shows greater facilitation for language acquisition than scripted speech data."

\noindent Output:
\begin{lstlisting}[style=jsonprompt]
{
  "Action": "use",
  "Agent": ["we"],
  "Object": {
    "Primary Object": ["the TILT-based cross lingual transfer learning approach"],
    "Secondary Object": ["<NONE>"]
  },
  "Context": ["<NONE>"],
  "Purpose": ["To examine the impact of native CDS"],
  "Method": ["<NONE>"],
  "Results": ["as in human SLA, language family distance predicts more negative transfer", "conversational speech data shows greater facilitation for language acquisition than scripted speech data"],
  "Analysis": ["<NONE>"],
  "Challenge": ["<NONE>"],
  "Ethical": ["<NONE>"],
  "Implications": ["<NONE>"],
  "Contradictions": ["<NONE>"]
}
\end{lstlisting}

\noindent \textbf{Conclusion Event}\\
For abstract: "Our findings call for further research using our novel Transformer-based SLA models and we would like to encourage it by releasing our code, data, and models."

\noindent Output:
\begin{lstlisting}[style=jsonprompt]
{
  "Action": "call for",
  "Agent": ["Our findings"],
  "Object": {
    "Primary Object": ["further research"],
    "Secondary Object": ["<NONE>"]
  },
  "Context": ["<NONE>"],
  "Purpose": ["<NONE>"],
  "Method": ["using our novel Transformer-based SLA models"],
  "Results": ["<NONE>"],
  "Analysis": ["<NONE>"],
  "Challenge": ["<NONE>"],
  "Ethical": ["<NONE>"],
  "Implications": ["we would like to encourage it by releasing our code, data, and models"],
  "Contradictions": ["<NONE>"]
}
\end{lstlisting}

\noindent \textbf{\# \# \# Output Format:}
\begin{lstlisting}[style=jsonprompt]
{
  "Action": "EXACT TEXT or <NONE>",
  "Agent": ["EXACT TEXT or <NONE>"],
   "Object": {
    "Primary Object": ["EXACT TEXT or <NONE>"],
    "Secondary Object": ["EXACT TEXT or <NONE>"]
  },
  "Context": ["EXACT TEXT or <NONE>"],
  "Purpose": ["EXACT TEXT or <NONE>"],
  "Method": ["EXACT TEXT or <NONE>"],
  "Results": ["EXACT TEXT or <NONE>"],
  "Analysis": ["EXACT TEXT or <NONE>"],
  "Challenge": ["EXACT TEXT or <NONE>"],
  "Ethical": ["EXACT TEXT or <NONE>"],
  "Implications": ["EXACT TEXT or <NONE>"],
  "Contradictions": ["EXACT TEXT or <NONE>"]
}
\end{lstlisting}

\noindent \textbf{\# \# \# IMPORTANT INSTRUCTIONS:}
\begin{itemize}
\item You MUST return ONLY ONE JSON structure.
\item NO explanation text, thinking, or commentary before or after the JSON.
\item NEVER repeat the JSON structure.
\item ALL fields must use arrays with [\texttt{"<NONE>"}] for missing arguments.
\item Follow the EXACT format shown in the template.
\item ONLY extract arguments that are explicitly present in the text. DO NOT hallucinate or add any information not found in the abstract.
\item Carefully study the one-shot examples to understand how arguments should be correctly annotated from the text.
\end{itemize}

\noindent \textbf{\# \# \# Output (JSON only)}
\end{tcolorbox}

\clearpage 
\begin{tcolorbox}[
    enhanced,
    breakable,
    width=\textwidth,
    enlarge left by=0mm,
    enlarge right by=0mm,
    title={True Event-Type Trigger Identification \& Argument Extraction Prompt},
    colback=white,
    colframe=black,
    fonttitle=\bfseries
]
You are an expert argument annotator. Given a part of the text and the event type from the scientific abstract (e.g., "Background", "Method", "Result", "Conclusion"), you need to identify the key trigger for the event (the main verb or action that signals an important research activity) and annotate the abstract with the corresponding argument components related to this trigger. Extractions should capture complete phrases around this key trigger and be organized in a single JSON format, containing only what is explicitly stated in the text without adding any interpretation.

\noindent \textbf{\# \# \# Event Type Definitions:}
\begin{itemize}
\item \textbf{Background}: Problem, motivation, context, research gap, or objectives.
\item \textbf{Method}: Techniques, experimental setups, frameworks, datasets.
\item \textbf{Result}: Main findings, discoveries, statistics, or trends.
\item \textbf{Conclusion}: Importance, impact, applications, or future work.
\end{itemize}

\noindent \textbf{\# \# \# \{event\_type\} Event Abstract Segment to Analyze: \# \# \#}

\noindent \{abstract\}

\noindent \textbf{\# \# \# Argument Components to Extract:}

\noindent \textbf{Action:} What is the SINGLE most representative trigger (verb or verb phrase) in the segment? 

\noindent \textbf{Agent:} Who or what is performing the Action? 

\noindent \textbf{Object:}
\begin{itemize}
\item \textbf{Primary Object:} What is directly receiving or affected by the Action? 
\item \textbf{Secondary Object:} What is a secondary entity also receiving the Action?
\end{itemize}

\noindent \textbf{Context:} What provides foundational or situational information of the event?

\noindent \textbf{Purpose:} What is the purpose or aim of the event?

\noindent \textbf{Method:} What techniques, tools, approaches, or frameworks are used in the event?

\noindent \textbf{Results:} What are the outcomes, observations or findings of the event?

\noindent \textbf{Analysis:} What are the interpretations or explanations of other arguments?

\noindent \textbf{Challenge:} What are the constraints or weaknesses of the event?

\noindent \textbf{Ethical:} What are the ethical concerns, justifications or implications of the event?

\noindent \textbf{Implications:} What is the broader significance or potential for future applications/research?

\noindent \textbf{Contradictions:} What are the disagreements with existing knowledge?

\noindent \textbf{\# \# \# Extraction Rules:}

\begin{enumerate}
\item Extract complete phrases, not just single words.
\item Only extract elements that are explicitly present. Mark missing elements as [\texttt{"<NONE>"}].
\item Use the exact text from the abstract.
\item Break down sentences when different parts fit different arguments.
\item NEVER use the same span of text for multiple arguments - each piece of text must be assigned to exactly one argument type. However, multiple text spans can be part of the same argument (e.g., [\texttt{"text span 1"}, \texttt{"text span 2".....}] can be used for a single argument type) if different parts of the text contribute to the same argument.
\item If text could fit multiple arguments, prioritize in this order: Results > Purpose > Method > Analysis > Implication > Challenge > Contradiction > Context > Ethical
\item Consider the event type when determining the most appropriate argument assignments.
\end{enumerate}

\noindent \textbf{\# \# \# Output Format:}
\begin{lstlisting}[style=jsonprompt]
{
  "Action": "EXACT TEXT or <NONE>",
  "Agent": ["EXACT TEXT or <NONE>"],
   "Object": {
    "Primary Object": ["EXACT TEXT or <NONE>"],
    "Secondary Object": ["EXACT TEXT or <NONE>"]
  },
  "Context": ["EXACT TEXT or <NONE>"],
  "Purpose": ["EXACT TEXT or <NONE>"],
  "Method": ["EXACT TEXT or <NONE>"],
  "Results": ["EXACT TEXT or <NONE>"],
  "Analysis": ["EXACT TEXT or <NONE>"],
  "Challenge": ["EXACT TEXT or <NONE>"],
  "Ethical": ["EXACT TEXT or <NONE>"],
  "Implications": ["EXACT TEXT or <NONE>"],
  "Contradictions": ["EXACT TEXT or <NONE>"]
}
\end{lstlisting}

\noindent \textbf{\# \# \# IMPORTANT INSTRUCTIONS:}
\begin{itemize}
\item You MUST return ONLY ONE JSON structure.
\item NO explanation text, thinking, or commentary before or after the JSON.
\item NEVER repeat the JSON structure.
\item ALL fields must use arrays with [\texttt{"<NONE>"}] for missing arguments.
\item Follow the EXACT format shown in the template.
\item ONLY extract arguments that are explicitly present in the text. DO NOT hallucinate or add any information not found in the abstract.
\item Use the provided event type to guide your analysis and ensure the extraction is appropriate for that type of event.
\end{itemize}

\noindent \textbf{\# \# \# Output (JSON only)}
\end{tcolorbox}

\clearpage

\begin{tcolorbox}[
    enhanced,
    breakable,
    width=\textwidth,
    enlarge left by=0mm,
    enlarge right by=0mm,
    title={Predict Event-Type Trigger Identification \& Argument Extraction Prompt},
    colback=white,
    colframe=black,
    fonttitle=\bfseries
]
You are an expert argument annotator. Given a part of the text from a scientific abstract, you need to first determine what type of event this text represents, then identify the key trigger for the event (the main verb or action that signals an important research activity) and annotate the abstract with the corresponding argument components related to this trigger. Based on the event type you determine, perform the argument extraction accordingly. Extractions should capture complete phrases around this key trigger and be organized in a single JSON format, containing only what is explicitly stated in the text without adding any interpretation.

\noindent \textbf{\# \# \# Event Type Definitions:}
\begin{itemize}
\item \textbf{Background}: Problem, motivation, context, research gap, or objectives.
\item \textbf{Method}: Techniques, experimental setups, frameworks, datasets.
\item \textbf{Result}: Main findings, discoveries, statistics, or trends.
\item \textbf{Conclusion}: Importance, impact, applications, or future work.
\end{itemize}

\noindent \textbf{\# \# \# Abstract Segment to Analyze: \# \# \#}

\noindent \{abstract\}

\noindent \textbf{\# \# \# Argument Components to Extract:}

\noindent \textbf{Action:} What is the SINGLE most representative trigger (verb or verb phrase) in the segment? 

\noindent \textbf{Agent:} Who or what is performing the Action? 

\noindent \textbf{Object:}
\begin{itemize}
\item \textbf{Primary Object:} What is directly receiving or affected by the Action? 
\item \textbf{Secondary Object:} What is a secondary entity also receiving the Action?
\end{itemize}

\noindent \textbf{Context:} What provides foundational or situational information of the event?

\noindent \textbf{Purpose:} What is the purpose or aim of the event?

\noindent \textbf{Method:} What techniques, tools, approaches, or frameworks are used in the event?

\noindent \textbf{Results:} What are the outcomes, observations or findings of the event?

\noindent \textbf{Analysis:} What are the interpretations or explanations of other arguments?

\noindent \textbf{Challenge:} What are the constraints or weaknesses of the event?

\noindent \textbf{Ethical:} What are the ethical concerns, justifications or implications of the event?

\noindent \textbf{Implications:} What is the broader significance or potential for future applications/research?

\noindent \textbf{Contradictions:} What are the disagreements with existing knowledge?

\noindent \textbf{\# \# \# Extraction Rules:}

\begin{enumerate}
\item Extract complete phrases, not just single words.
\item Only extract elements that are explicitly present. Mark missing elements as [\texttt{"<NONE>"}].
\item Use the exact text from the abstract.
\item Break down sentences when different parts fit different arguments.
\item NEVER use the same span of text for multiple arguments - each piece of text must be assigned to exactly one argument type. However, multiple text spans can be part of the same argument (e.g., [\texttt{"text span 1"}, \texttt{"text span 2".....}] can be used for a single argument type) if different parts of the text contribute to the same argument.
\item If text could fit multiple arguments, prioritize in this order: Results > Purpose > Method > Analysis > Implication > Challenge > Contradiction > Context > Ethical
\item Consider the event type when determining the most appropriate argument assignments.
\end{enumerate}

\noindent \textbf{\# \# \# Output Format:}
\begin{lstlisting}[style=jsonprompt]
{
  "Action": "EXACT TEXT or <NONE>",
  "Agent": ["EXACT TEXT or <NONE>"],
   "Object": {
    "Primary Object": ["EXACT TEXT or <NONE>"],
    "Secondary Object": ["EXACT TEXT or <NONE>"]
  },
  "Context": ["EXACT TEXT or <NONE>"],
  "Purpose": ["EXACT TEXT or <NONE>"],
  "Method": ["EXACT TEXT or <NONE>"],
  "Results": ["EXACT TEXT or <NONE>"],
  "Analysis": ["EXACT TEXT or <NONE>"],
  "Challenge": ["EXACT TEXT or <NONE>"],
  "Ethical": ["EXACT TEXT or <NONE>"],
  "Implications": ["EXACT TEXT or <NONE>"],
  "Contradictions": ["EXACT TEXT or <NONE>"]
}
\end{lstlisting}

\noindent \textbf{\# \# \# IMPORTANT INSTRUCTIONS:}
\begin{itemize}
\item You MUST return ONLY ONE JSON structure.
\item NO explanation text, thinking, or commentary before or after the JSON.
\item NEVER repeat the JSON structure.
\item ALL fields must use arrays with [\texttt{"<NONE>"}] for missing arguments.
\item Follow the EXACT format shown in the template.
\item ONLY extract arguments that are explicitly present in the text. DO NOT hallucinate or add any information not found in the abstract.
\item Use the provided event type to guide your analysis and ensure the extraction is appropriate for that type of event.
\end{itemize}

\noindent \textbf{\# \# \# Output (JSON only)}
\end{tcolorbox}

\clearpage
\twocolumn

\section{Argument Extraction with EM Metrics and detailed Human Performance Comparison}
\label{app:Supplementary Experiment Results}
\begin{table}[t]
\centering
\small
\renewcommand{\arraystretch}{1.2}
\setlength{\tabcolsep}{1.5pt}
\begin{tabular}{@{}lcccccc@{}}
\hline
\multirow{2}{*}{\textbf{Methods}} & \multicolumn{3}{c}{\textbf{Arg-I (EM)}} & \multicolumn{3}{c}{\textbf{Arg-C (EM)}} \\
\cline{2-7}
& \textbf{P} & \textbf{R} & \textbf{F1} & \textbf{P} & \textbf{R} & \textbf{F1} \\
\hline
\textit{Tuning-based models}\\
EEQA & 14.26 & 15.01 & 14.63 & 11.59 & 12.20 & 11.88 \\
DEGREE & \textbf{44.97} & 12.69 & 19.79 & \textbf{34.23} & 9.66 & 15.07 \\
OneIE & 32.03 & \textbf{35.2}7 & \textbf{33.57} & 25.38 & \textbf{27.95} & \textbf{26.61} \\
\hline
\textit{Zero-shot LLMs}\\
DS-R1-Llama & 10.33 & 5.46 & 7.15 & 6.23 & 3.30 & 4.31 \\
Qwen & 9.59 & 7.10 & 8.16 & 5.08 & 3.76 & 4.33 \\
Llama & 7.01 & 7.17 & 7.09 & 3.73 & 3.81 & 3.77 \\
GPT & 17.84 & 23.03 & 20.10 & 13.37 & 17.27 & 15.07 \\
\hline
\textit{One-shot LLMs}\\
DS-R1-Llama & 13.28 & 5.51 & 7.79 & 7.08 & 2.93 & 4.15 \\
Qwen & 13.98 & 9.16 & 11.07 & 7.24 & 4.74 & 5.73 \\
Llama & 13.02 & 9.93 & 11.27 & 6.55 & 5.00 & 5.67 \\
GPT & 25.75 & 25.79 & 25.77 & 19.38 & 19.41 & 19.4 \\
\hline
\end{tabular}
\caption{EM-based Precision (P), Recall (R), and F1-score (\%) on baseline models for argument identification (Arg-I) and classification (Arg-C) tasks.}
\label{tab:EM_Results_Overall}
\vspace{-5mm}
\end{table}

% \begin{table}[t]
% \centering
% \small
% \renewcommand{\arraystretch}{1.2}
% \setlength{\tabcolsep}{1.5pt}
% \begin{tabular}{@{}lcccccc@{}}
% % {|l|ccc|ccc|} {@{}lccccccl@{}}
% \hline
% \multirow{2}{*}{\textbf{Methods}} & \multicolumn{3}{c}{\textbf{Arg-I (IoU)}} & \multicolumn{3}{c}{\textbf{Arg-C (IoU)}} \\
% \cline{2-7}
% & \textbf{P} & \textbf{R} & \textbf{F1} & \textbf{P} & \textbf{R} & \textbf{F1} \\
% % \hline
% % Human & 67.52 & 52.46 & 59.04 & 52.33 & 40.65 & 45.76 \\
% \hline
% \textit{Tuning-based models}\\
% EEQA & 32.09 & 33.77 & 32.91 & 25.85 & 27.20 & 26.51 \\
% DEGREE & \textbf{67.79} & 19.13 & 29.84 & \textbf{48.99} & 13.83 & 21.57 \\
% OneIE & 51.11 & \textbf{56.29} & \textbf{53.57} & 39.69 & \textbf{43.71} & \textbf{41.61} \\
% \hline
% \textit{Zero-shot LLMs}\\
% DeepSeek-R1 & 31.11 & 16.46 & 21.53 & 16.32 & 8.63 & 11.29 \\
% Qwen & 35.68 & 26.41 & 30.35 & 17.58 & 13.01 & 14.96 \\
% Llama & 24.37 & 24.90 & 24.63 & 11.68 & 11.93 & 11.80 \\
% GPT & 43.03 & 55.56 & 48.50 & 30.40 & 39.25 & 34.26 \\
% \hline
% \textit{One-shot LLMs}\\
% DeepSeek-R1 & 42.62 & 17.67 & 24.98 & 19.59 & 8.12 & 11.48 \\
% Qwen & 46.33 & 30.36 & 36.69 & 20.96 & 13.74 & 16.60 \\
% Llama & 44.70 & 34.08 & 38.68 & 18.93 & 14.44 & 16.38 \\
% GPT & 50.14 & 50.22 & 50.18 & 34.60 & 34.66 & 34.63 \\
% \hline
% \end{tabular}
% \caption{IoU-based Precision (P), Recall (R), and F1-score (\%) on baseline models for argument identification (Arg-I) and classification (Arg-C) tasks.}
% \label{tab:IoU_Results_Overall}
% \vspace{-5mm}
% \end{table}

\begin{figure}[h]
    \centering
    \includegraphics[width=0.98\linewidth]{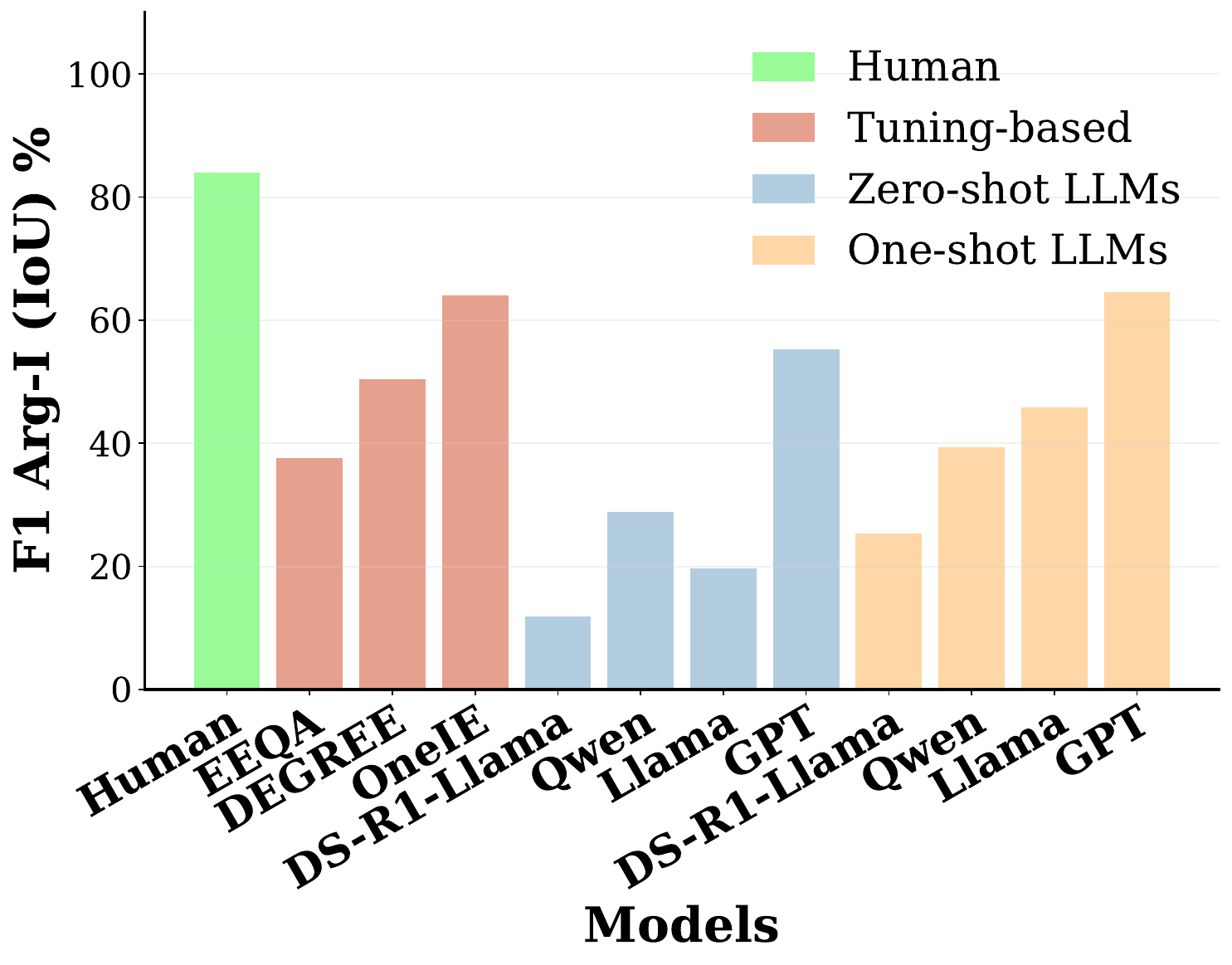}
    \caption{Performance comparison of various methods on argument identification (Arg-I) using IoU F1 scores. Methods are grouped by type: Human baseline, tuning-based models, zero-shot LLMs, and one-shot LLMs.}
    \label{fig:arg_i_performance}
    \vspace{-3mm}
\end{figure}

\begin{figure}[h]
    \centering
    \includegraphics[width=0.98\linewidth]{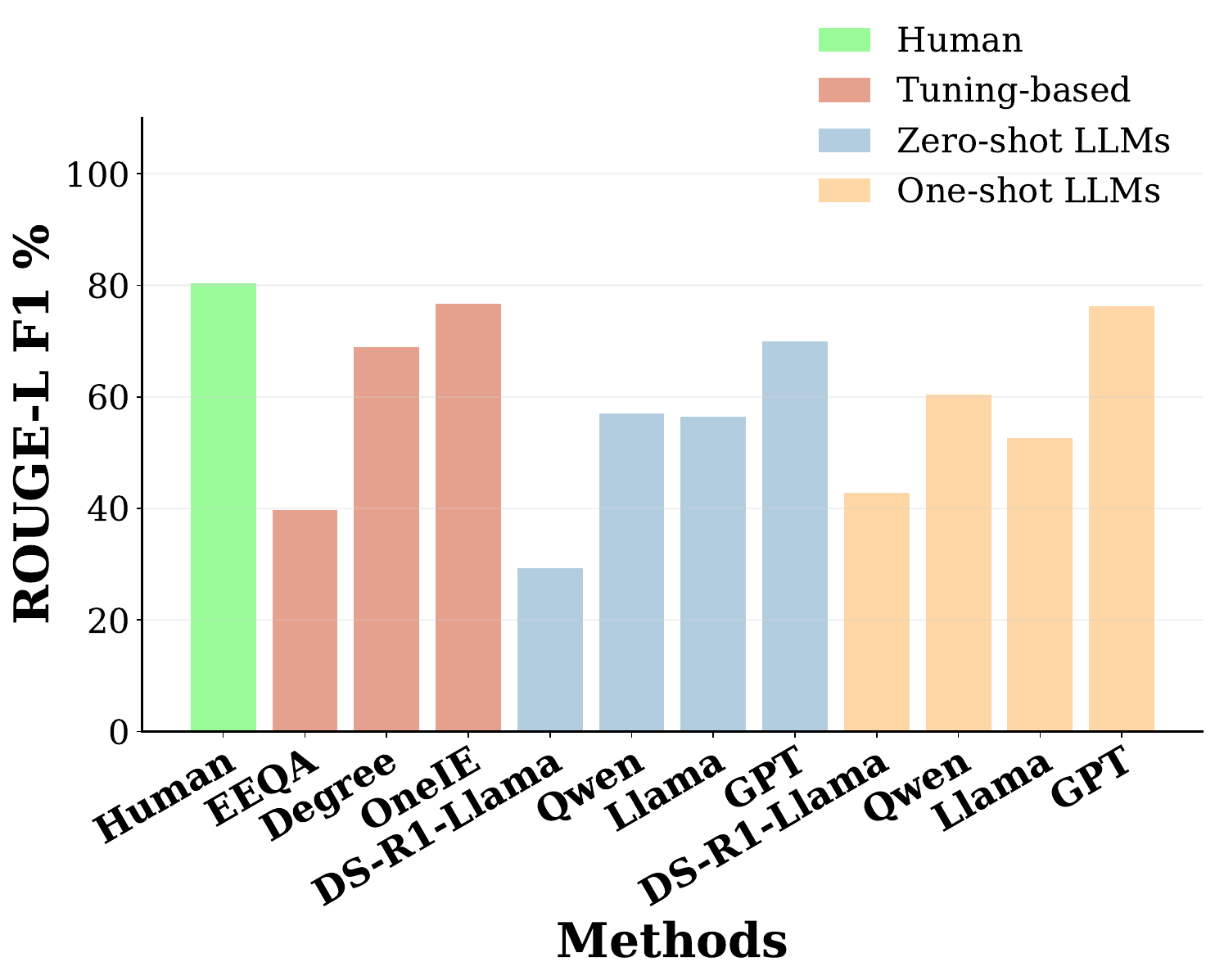}
    \caption{Performance comparison of various methods on ROUGE-L F1 scores. Methods are grouped by type: Human baseline, tuning-based models, zero-shot LLMs, and one-shot LLMs.}
    \label{fig:human rouge-L}
\end{figure}

In Section~\ref{sec:results}, we analyze argument extraction under the IoU metric and examined the human–model performance gap for argument classification (Arg-C) using IoU. Here, we complement that analysis by reporting results under the EM metric for argument extraction, as well as argument identification (Arg-I) and trigger identification with ROUGE-L human–model gaps, to provide a more comprehensive evaluation. As shown in Table~\ref{tab:EM_Results_Overall}, OneIE remains the best-performing model, while DEGREE continues to exhibit high precision but low recall. Among LLMs, GPT-4.1 consistently achieves the best performance, and one-shot prompting again improves results across all LLMs. Overall, the findings remain consistent---switching from IoU to EM does not alter the relative comparison between models, but EM results in lower scores for all models due to its stricter matching criteria.

Figure~\ref{fig:arg_i_performance} shows the Arg-I performance gap between humans and models, which closely mirrors the Arg-C results. The gap remains around 20\%, highlighting the need for multi-domain scientific EE models. In contrast, Figure~\ref{fig:human rouge-L} reveals a smaller gap in ROUGE-L scores for trigger identification, indicating that this task is considerably easier and most models perform well. Nevertheless, since argument extraction is the core challenge, there remains significant room for improvement in addressing multi-domain scientific EE.

\section{Effects of removal of domains on each tuning-based model}
\label{app:RQ3}
\begin{figure}[h]
    \centering
    \includegraphics[width=0.98\linewidth]{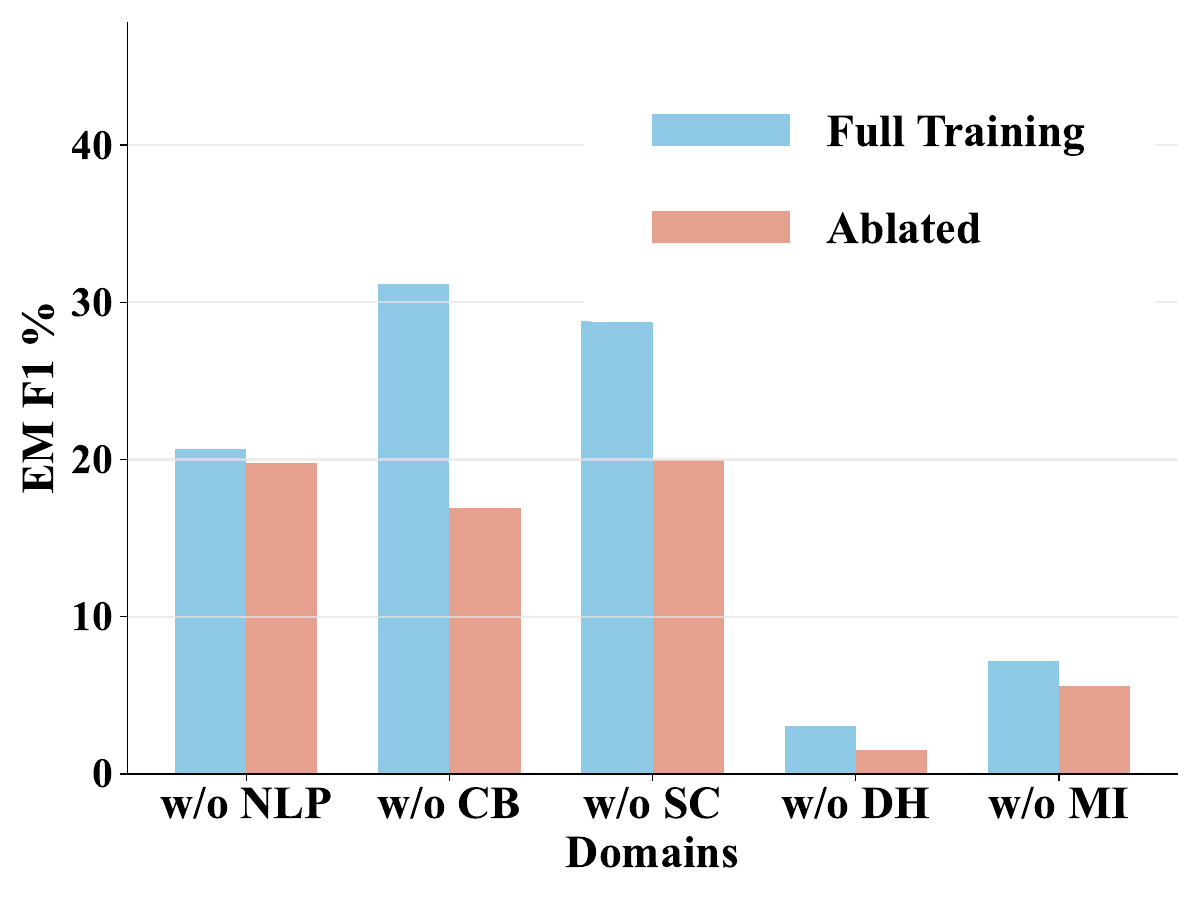}
    \caption{F1-scores reported for full training versus training with one domain removed for DEGREE under Exact Match (EM).}
    \label{fig:degree_em_performance}
\end{figure}

\begin{figure}[h]
    \centering
    \includegraphics[width=0.98\linewidth]{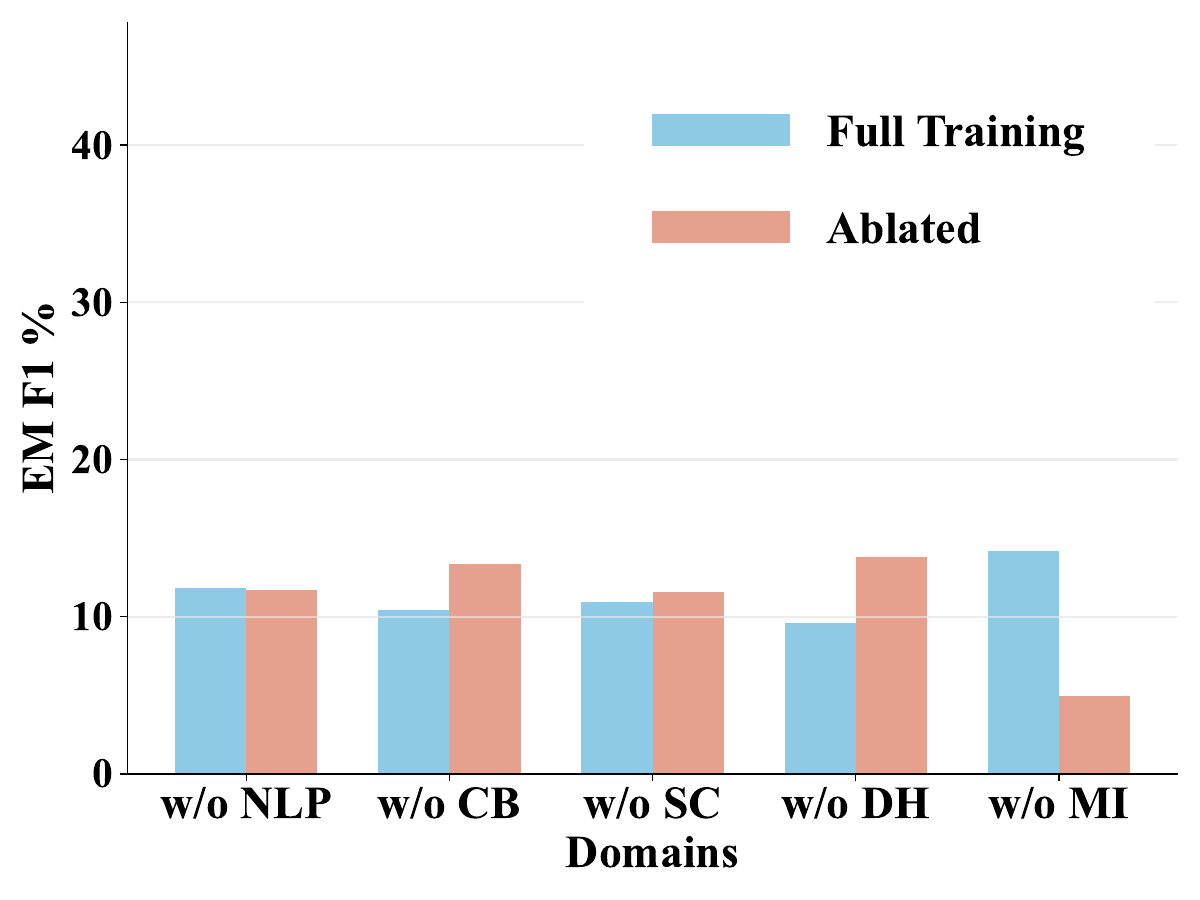}
    \caption{F1-scores reported for full training versus training with one domain removed for EEQA under Exact Match (EM).}
    \label{fig:eeqa_em_performance}
\end{figure}

We present domain ablation results for DEGREE and EEQA under the EM setting in Figure~\ref{fig:degree_em_performance} and Figure~\ref{fig:eeqa_em_performance}, respectively. For DEGREE, removing a domain consistently leads to performance drops, similar to OneIE, though the impact is generally smaller. This suggests DEGREE benefits from domain-specific training but is somewhat more resilient, possibly due to its generative nature. In contrast, EEQA shows minimal sensitivity to domain removal. This may be because its QA-based design relies more on question formulation and span selection, making it less dependent on domain-specific linguistic patterns.

\section{Trigger and Argument Identification by Event Types and Domains}
\label{app:RQ1 RQ2}
Results for trigger identification and argument identification are presented by event type and domain, providing supplementary detail to the analysis in Section~\ref{sec:results} and offering deeper insight into how event types and domains impact SciEvent performance.

\begin{figure*}[h]
    \centering
    \vspace{-5mm}
    \includegraphics[height=0.23\textheight, width=0.98\linewidth]{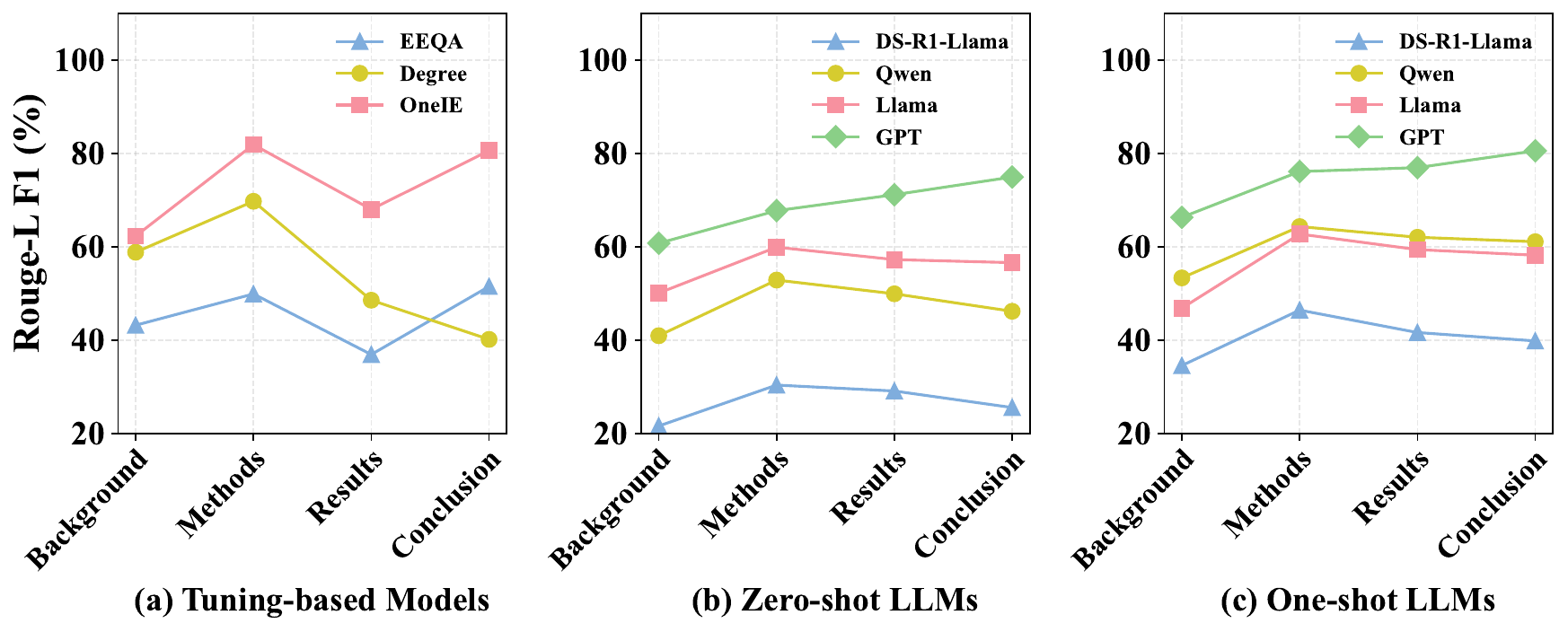}
    \vspace{-3mm}
    \caption{Comparison of Rouge-L F1 scores (\%) across different event types.}
    \label{fig:Rouge-L event type}
\end{figure*}

\begin{figure*}[h]
    \centering
    \vspace{-5mm}
    \includegraphics[height=0.23\textheight, width=0.98\linewidth]{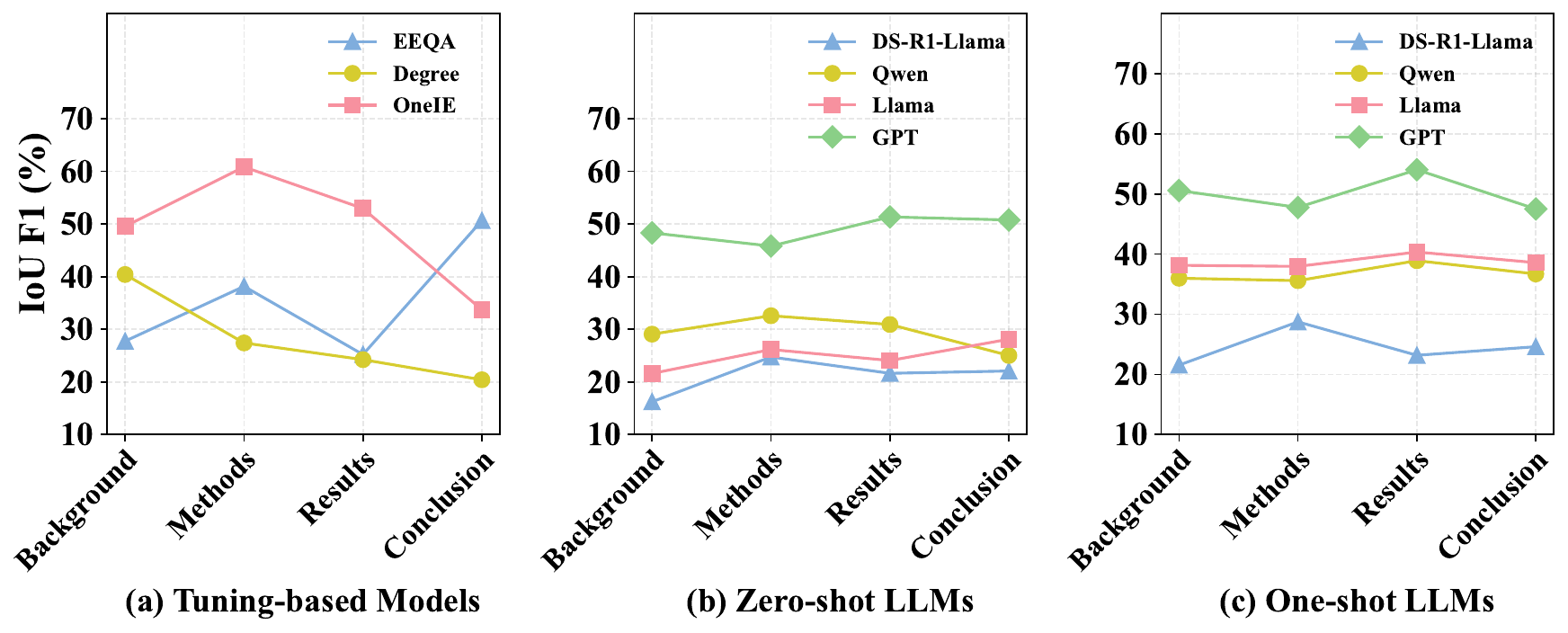}
    \vspace{-3mm}
    \caption{Comparison of Intersection-over-Union (IoU) on Arg-I F1-scores (\%) across different event types.}
    \label{fig:arg_I_event_type_wise_iou_comparison}
\end{figure*}

Figure~\ref{fig:Rouge-L event type} presents ROUGE-L scores for trigger identification by event type, where the Conclusion event achieves the highest performance. This is likely due to its shorter and simpler structure, offering fewer candidate verbs, making trigger extraction easier. The performance trends for other event types are similar to those discussed in Sections~\ref{sec:results} on argument classification. Figure~\ref{fig:arg_I_event_type_wise_iou_comparison} reports IoU scores for argument identification, which closely mirror the argument classification results but show an overall performance increase of about 20\%, due to the easier argument identification task.

\begin{figure*}[h]
    \centering
    \vspace{-5mm}
    \includegraphics[height=0.23\textheight, width=0.98\linewidth]{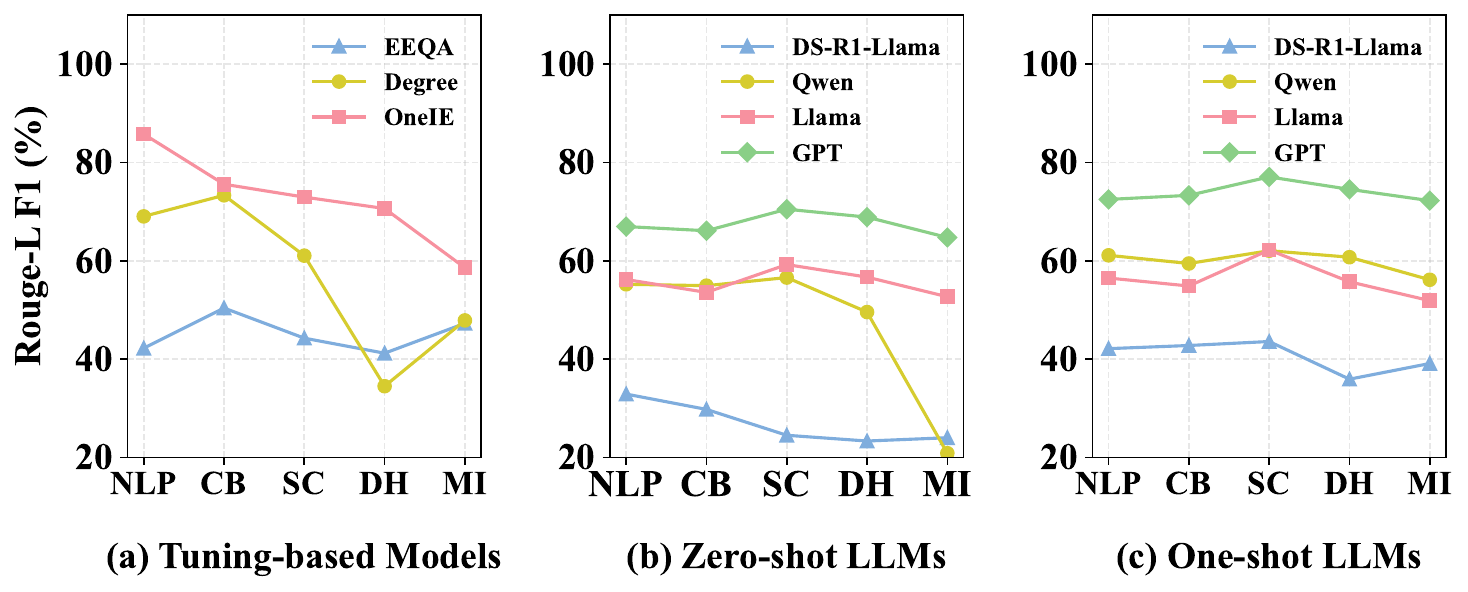}
    \vspace{-3mm}
    \caption{Comparison of Rouge-L F1 scores (\%) across different academic domains.}
    \label{fig:Rouge-L domain}
\end{figure*}

\begin{figure*}[h]
    \centering
    \vspace{-5mm}
    \includegraphics[height=0.23\textheight, width=0.98\linewidth]{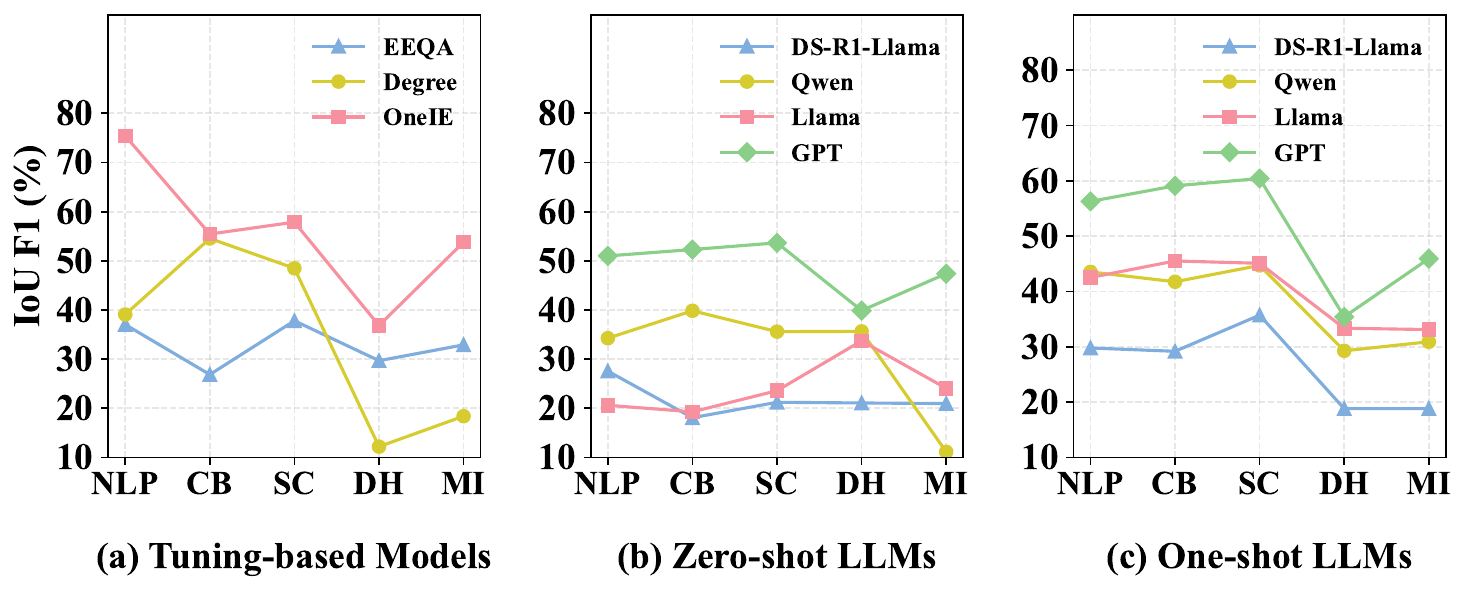}
    \vspace{-3mm}
    \caption{Comparison of Intersection-over-Union (IoU) on Arg-I F1-scores (\%) across different academic domains.}
    \label{fig:arg_I_venue_wise_iou_comparison}
\end{figure*}

Figure~\ref{fig:Rouge-L domain} shows some difference in the Medical Informatics (MI) domain compared to argument classification. MI exhibits lower trigger identification performance, due to longer texts containing more verbs, which increases ambiguity and makes trigger extraction more difficult.
Figure~\ref{fig:arg_I_venue_wise_iou_comparison} again shows a 20\% performance boost across all models, due to the easier argument identification task, while preserving trends consistent with those observed in argument classification.
\section{SciEvent keywords analysis}
\label{app:keyword}
We present a detailed keyword analysis grounded in each domain’s call for papers in table \ref{tab:domain_counts}. For Digital Humanities (DHq 2021–-2023), we include the majority of abstracts from 2021 to 2023 due to limited publications, ensuring comprehensive coverage and minimizing bias. On the other hand, as shown in the tables below for the rest four domains, we observe that our dataset covers all major research topics outlined in each venue’s call for papers. This suggests that our benchmark includes a diverse set of scientific articles and is reasonably representative within each domain.
% EEQA EM Performance

\clearpage

\onecolumn  
\newpage
\section{Detailed example of SciEvent dataset}
\begin{figure}[h]
    \centering
    \includegraphics[width=0.98\linewidth]{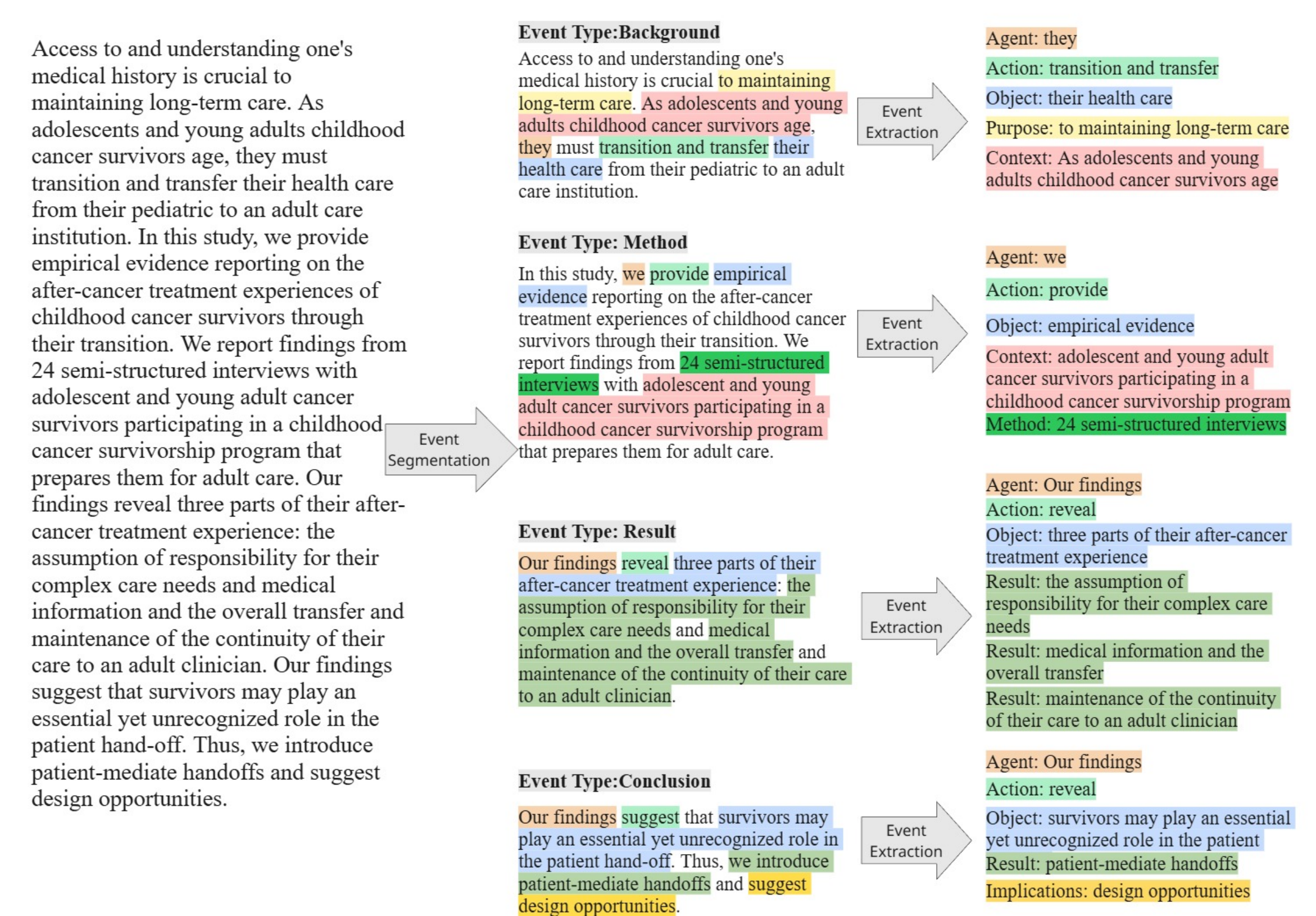}
    \caption{Full event extraction example from SciEvent, including event segmentation and event extraction, where trigger is a tuple including Agent, Action and Object.}
    \label{fig: SciEvent Example}
\end{figure}

We show one detailed example of SciEvent dataset, including event segmentation and event extraction in Figrue~\ref{fig: SciEvent Example}

\twocolumn
\begin{table}[t]
\centering
\small
\renewcommand{\arraystretch}{1.1}
\setlength{\tabcolsep}{4pt}
\begin{tabular}{@{}lr@{}}
\hline
\textbf{Domain} & \textbf{Count} \\
\hline
\textit{NLP (ACL 2023)}\\
Computational Social Science and Cultural Analytics & 2 \\
Dialogue and Interactive Systems & 10 \\
Discourse and Pragmatics & 2 \\
Ethics and NLP & 8 \\
Generation & 26 \\
Information Extraction & 10 \\
Information Retrieval and Text Mining & 2 \\
Interpretability and Analysis of Models for NLP & 24 \\
Language Grounding to Vision, Robotics and Beyond & 2 \\
Multilingualism and Language Contact & 16 \\
Linguistic Theories, Cognitive Modeling, and Psycholinguistics & 6 \\
Machine Learning for NLP & \textbf{50} \\
Machine Translation & 6 \\
NLP Applications & 3 \\
Phonology, Morphology, and Word Segmentation & 2 \\
Question Answering & 8 \\
Resources and Evaluation & \textbf{62} \\
Semantics: Lexical & 2 \\
Semantics: Sentence-level, Textual Inference, Other Areas & 4 \\
Sentiment, Stylistic, Argument Mining & 6 \\
Speech and Multimodality & 10 \\
Summarization & 6 \\
Syntax: Tagging, Chunking and Parsing & 6 \\
\hline
\textit{CB (Bioinformatics 2023)}\\
Genome analysis & 6 \\
Sequence analysis & 4 \\
Phylogenetics & 4 \\
Structural bioinformatics & 14 \\
Gene expression & 16 \\
Genetic and population analysis & \textbf{18} \\
Systems biology & 14 \\
Data and text mining & 6 \\
Databases and ontologies & 12 \\
Bioimage informatics & 4 \\
\hline
\textit{SC (CSCW 2023)}\\
Social and crowd computing & 67 \\
System development & 6 \\
Theory & 42 \\
Empirical investigations & \textbf{78} \\
Data mining and modeling & 27 \\
Methodologies and tools & \textbf{77} \\
Domain-specific social and collaborative applications & 31 \\
Collaboration systems based on emerging technologies & 7 \\
Ethics and policy implications & 33 \\
Crossing boundaries & 19 \\
\hline
\textit{MI (JMIR 2023)}\\
Clinical Decision Support & \textbf{47} \\
Automated Feedback & 7 \\
Virtual Patient Development & 25 \\
Content Quality and Prompting & 23 \\
AI Curriculum Design & 11 \\
Patient Education via ChatGPT & 19 \\
Preparing for AI-Literate Patients & 9 \\
Ethics and Legal Concerns & \textbf{44} \\
Academic Integrity and Policy & 15 \\
Trends and Use Cases & 24 \\
Future Outlook & 3 \\
Practical Tutorials & 8 \\
\hline
\end{tabular}
\caption{Domain distribution and counts across different research venues and conferences.}
\label{tab:domain_counts}
\vspace{-3mm}
\end{table}

\end{document}